\documentclass[journal]{IEEEtran}
\usepackage{graphicx}
\usepackage{graphics} 
\usepackage{epsfig} 
\usepackage{times} 
\usepackage{amsmath} 
\usepackage{amssymb}  
\usepackage{enumerate}
\usepackage{mathrsfs}
\usepackage{multirow}
\usepackage{subfigure}
\usepackage{latexsym}
\usepackage{bm}
\usepackage{cases}
\usepackage{dsfont}
\usepackage{multicol}
\usepackage{multirow}
\usepackage{color}
\usepackage{url}
\usepackage{algorithmic}
\usepackage{algorithm}
\usepackage{comment}
\usepackage{textcomp}
\usepackage{booktabs}
\usepackage{cite}
\usepackage[hidelinks]{hyperref}
\usepackage{cases}
\usepackage{footmisc}
\usepackage{cases} 
\newtheorem{remark}{Remark}
\newtheorem{assumption}{Assumption}
\newtheorem{lemma}{Lemma}
\newtheorem{proposition}{Proposition}
\newtheorem{definition}{Definition}

\newtheorem{theorem}{Theorem}

{\renewcommand{\arraystretch}{2.0}
	\begin{bmatrix}}%
	{\end{bmatrix}
	\renewcommand{\arraystretch}{1.0}}

\def\build#1_#2^#3{\mathrel{\mathop{\kern0pt#1}\limits_{#2}^{#3}}}%
\def\argmin#1{\build{\rm argmin}_{#1}^{}}
\def\build#1_#2^#3{\mathrel{\mathop{\kern0pt#1}\limits_{#2}^{#3}}}%
\def\min#1{\build{\rm min}_{#1}^{}}
\def\build#1_#2^#3{\mathrel{\mathop{\kern0pt#1}\limits_{#2}^{#3}}}%

\def\hlinewd#1{%
	\noalign{\ifnum0=`}\fi\hrule \@height #1 %
	\futurelet\reserved@a\@xhline}
\ifCLASSINFOpdf
\else
\fi
\begin{document}
	%
	\title{Model-Based Safe Reinforcement Learning with Time-Varying State and Control Constraints: An Application to Intelligent Vehicles}
	
	
	\author{\IEEEauthorblockN{Xinglong Zhang~\IEEEmembership{Member,~IEEE},
			Yaoqian Peng,
			Biao Luo~\IEEEmembership{Senior member,~IEEE}, 
			Wei Pan~\IEEEmembership{Member,~IEEE}, 
			Xin Xu~\IEEEmembership{Senior member,~IEEE}, and
			Haibin Xie} 
		
		\thanks{
			This work has been submitted to the IEEE for possible publication. Copyright may be transferred without notice, after which this version may no longer be accessible.			
			Xinglong Zhang, Yaoqian Peng,  Xin Xu, and Haibin Xie are with the College of Intelligence Science and Technology, National University of Defense Technology, Changsha, China. Biao Lu is with the School of Automation, Central South University, Changsha, China.
			Wei Pan is with the Department of Cognitive Robotics, Delft University of Technology, the Netherlands.
			Corresponding to: xinxu@nudt.edu.cn, zhangxinglong18@nudt.edu.cn.}}

	\markboth{Journal of \LaTeX\ Class Files,~Vol.~14, No.~8, August~2023}%
	{Xinglong Zhang \MakeLowercase{\textit{et al.}}: Model-Based Safe Reinforcement Learning with Time-Varying State and Control Constraints: An Application to Intelligent Vehicles}
	%



	
	\IEEEtitleabstractindextext{%
		\begin{abstract}
		In recent years, safe reinforcement learning (RL) with the actor-critic structure has gained significant interest for continuous control tasks. However, achieving near-optimal control policies with safety and convergence guarantees remains challenging. Moreover, few works have focused on designing RL algorithms that handle time-varying safety constraints.
		This paper proposes a safe RL algorithm for optimal control of nonlinear systems with time-varying state and control constraints. The algorithm's novelty lies in two key aspects. Firstly, the approach introduces a unique barrier force-based control policy structure to ensure control safety during learning. Secondly, a multi-step policy evaluation mechanism is employed, enabling the prediction of policy safety risks under time-varying constraints and guiding safe updates. Theoretical results on learning convergence, stability, and robustness are proven.
		The proposed algorithm outperforms several state-of-the-art RL algorithms in the simulated Safety Gym environment. It is also applied to the real-world problem of integrated path following and collision avoidance for two intelligent vehicles – a differential-drive vehicle and an Ackermann-drive one. The experimental results demonstrate the impressive  sim-to-real transfer capability of our approach, while showcasing satisfactory online control performance.
		\end{abstract}

		\begin{IEEEkeywords}
			Safe reinforcement learning, time-varying constraints, barrier force, multi-step policy evaluation.
	\end{IEEEkeywords}}
	
	\maketitle
	
	\IEEEdisplaynontitleabstractindextext

	%
	\IEEEpeerreviewmaketitle
	
	\section{Introduction}
	Reinforcement learning (RL) is promising for solving nonlinear optimal control problems and has received significant attention in the past decades, see~\cite{grondman2012survey,kim2021dynamic} and the references therein. 
	Until recently, significant progress has been made on RL with the actor-critic structure for continuous control tasks~\cite{grondman2012survey,haarnoja2018soft,liu2013policy,zhang2009neural,jiang2012computational,lim2020prediction,wang2021intelligent,8936917}. 
	In actor-critic RL, the value function and control policy are represented by the critic and actor networks, respectively, and learned via extensive policy exploration and exploitation. 
	However, the resulting learning-based control system might not guarantee safety for systems with state and stability constraints. {\color{black}It is known that safety constraint satisfaction is crucial besides optimality in many real-world robot control applications~\cite{garcia2015comprehensive,chow2019lyapunov}.} {\color{black}For instance, autonomous driving has been viewed as a promising technology that will bring fundamental changes to everyday life. 
		Still, one of the crucial issues concerns how to learn to drive safely under dynamic and unknown environments with unexpected obstacles~\cite{Kesslermixed2022}. For these practical reasons, many safe RL algorithms have been recently developed for safety-critical systems, see e.g.~\cite{chow2019lyapunov,berkenkamp2018safe,garcia2015comprehensive,srinivasan2020learning,yu2019convergent,xu2021crpo,ma2021model,yang2021accelerating,huh2020safe,zheng2021safe,marvi2021safe,chen2021context,li2021safe,brunke2021safe,richards2018lyapunov} 
		and the references therein.  Note that there are fruitful works in adaptive control with constraints, but the technique used differs from that in RL, for related references in adaptive control might refer to~\cite{sun2016neural,kong2019adaptive}.}   
	
	{\color{black}
		In general, current safe RL solutions can be categorized into three main approaches. 
		(i) The first family utilizes a unique mechanism in the learning procedure for safe policy optimization using, e.g., control barrier functions~\cite{yang2019safety,ma2021model,marvi2021safe}, formal verification~\cite{fulton2018safe,turchetta2020safe}, shielding~\cite{alshiekh2018safe,zanon2020safe,thananjeyan2021recovery}, and external intervention~\cite{saunders2017trial,wagener2021safe}. These methods are prone to safe-biased learning by sacrificing greatly on performance. 
		And some of them rely on extra-human interference~\cite{saunders2017trial,wagener2021safe}.
		(ii) The second family proposes safe RL algorithms via primal-dual methods~\cite{chen2016stochastic,paternain2019safe,xu2021crpo,ding2021provably}. 
		In the resulting optimization problem, the Lagrangian multiplier serves as an extra weight whose update is noted sensitive to the control performance~\cite{xu2021crpo}.  Moreover, some optimization problems such as those with ellipsoidal constraints (covered in this work) could not satisfy the strong duality condition~\cite{zhang2021robust}.
		(iii) The third is reward/cost shaping-based RL approaches~\cite{geibel2005risk,balakrishnan2019structured,hu2020learning,tessler2018reward} where the cost functions are augmented  with various safety-related parts, e.g., barrier functions. As stated in~\cite{paternain2019safe}, such a design only informs the goal of guaranteeing safety by minimizing the reshaped cost function but fails to guide how to achieve it well through an actor-critic structure design. Specifically, the reshaped cost function values, which are usually evaluated in discrete-time instants, might change rapidly when the system variables approach the constraint boundary. Consequently, the weights of actor and critic networks are prone to divergence in the training process.
		These issues motivated our barrier force-based (simplified as barrier-based) actor-critic structure. In this work, we incorporate this unique structure into a control-theory-based RL framework, where model-based multi-step policy evaluation mechanism is also utilized to ensure convergence and safety in online learning scenarios. Moreover, few works have addressed the safe RL algorithm design under time-varying safety constraints.}
	
	This work proposes a model-based safe RL algorithm with theoretical guarantees for optimal control with time-varying state and control constraints. A new barrier force-based control policy (BCP) structure is constructed in the proposed safe RL approach, generating repulsive control forces as states and controls move toward the constraint boundaries. 
	Moreover, the time-varying constraints are addressed by a multi-step policy evaluation (MPE).  The proposed safe RL approach is implemented by an online barrier force-based actor-critic learning algorithm.
	The closed-loop theoretical property of our approach under nominal and perturbed cases and the convergence condition of the barrier-based actor-critic (BAC) learning algorithm is derived.  The effectiveness of our approach is tested on both simulations and real-world intelligent vehicles.  The effectiveness of our approach is tested on both simulations and real-world intelligent vehicles. %
	Our contributions are summarized as follows.
	
			(i) We proposed a safe RL for optimal control under time-varying state and control constraints. Under certain conditions (see Sections~\ref{sec:32}-A and -B), safety can be guaranteed in both online and offline learning scenarios. The performance and advantages of the proposed approach are achieved by two novel designs. The first is a barrier force-based control policy to ensure safety with an actor-critic structure. The second is a multi-step evaluation mechanism to predict the control policy’s future influence on the value function under time-varying safety constraints and guide the policy to update safely.
			
			(ii) We proved that the proposed safe RL could guarantee stability and robustness in the nominal scenario and under external disturbances, respectively. Also, the convergence condition of the  actor-critic implementation was derived by the Lyapunov method.
			
			(iii) 
			The proposed approach was applied to solve an integrated path following and collision avoidance problem of intelligent vehicles so that the control performance can be optimized with theoretical guarantees even with external disturbances. a) Extensive simulation results illustrate that our approach outperforms other state-of-the-art safe RL methods in learning safety and performance. b) We verified our approach’s offline sim-to-real transfer capability and real-world online learning performance, as well as the strengths to state-of-the-art model predictive control (MPC) algorithms.
	
	The remainder of the paper is organized as follows. Section II introduces the considered control problem and preliminary solutions. Section III presents the proposed safe RL approach and the BAC learning algorithm, while Section IV presents the main theoretical results. Section V shows the real-world experimental results, while some conclusions are drawn in Section VI. Some proofs of the theoretical results and additional experimental results are given in the appendix. 
	
	\textbf{Notation:} We denote $\mathbb{N}$ and $\mathbb{N}_{a}^{b}$as the set of natural numbers and integers $a,a+1,\cdots,b$. For a vector $x\in\mathbb{R}^{n}$, we denote $\|x\|_Q^2$ as $x^{\top}Qx$ and $\|x\|$ as the Euclidean norm. For a function $f(x)$ with an argument $x$, we denote  $\triangledown f(x)$ as the gradient to $x$. For a function $f(x,u)$ with arguments $x$ and $u$, we denote $\triangledown_zf(x,u)$ as the partial gradient to $z$, $z=x$ or $u$. Given a matrix $A\in\mathbb{R}^{n\times n}$, we use $\lambda_{\rm min}(A)$ ($\lambda_{\rm max}(A)$) to denote the minimal (maximal) eigenvalues. 
	We denote $\rm Int(\mathcal{Z})$ as the interior of a general set $\mathcal{Z}$. 
	For variables $z_{i}\in{\mathbb{R}}^{q_{i}}$,
	$i\in\mathbb{N}_1^M$, we define $(z_{1}, z_{2}, \cdots, z_{\rm\scriptscriptstyle M})=[\,z_{1}^{\top}\ z_{2}^{\top}\ \cdots\ z_{\rm\scriptscriptstyle M}^{\top}\,]^{\top}\in{\mathbb{R}}^{q}$, where $q= \sum_{i=1}^{M}q_{i}$.
	\section{Problem Formulation}
	In this section, we first describe the considered model and the associated safety constraints. Then, the optimal control objective and preliminary RL results are given. Finally, the safe RL problem formulation by cost reconstruction with barrier functions is presented. 
	\subsection{System Model and Constraints}
	The considered system under control is a class of discrete-time nonlinear systems described by
	\begin{equation}\label{Eqn:LL}
	\begin{array}{ll}
	x_{k+1}=f(x_{k},u_{k}),
	\end{array}
	\end{equation}
	where $x_{k}\in\mathcal{X}_k\subseteq{\mathbb{R}}^{n}$ and $u_{k}\in\mathcal{U}_k\subseteq {\mathbb{R}}^{m}$
	are the state and input variables,  $k$ is the discrete-time index,  {\color{black}$\mathcal{X}_k=\{x\in\mathbb{R}^n|G^{i}_{x,k}(x)\leq0,\,\forall i\in\mathbb{N}_1^{p_x}\}$ and $\mathcal{U}_k=\{u\in\mathbb{R}^m|G^{i}_{u,k}(u)\leq0,\,\forall i\in\mathbb{N}_1^{p_u}\}$} are sets that represent time-varying constraints,  $\{0\}\subseteq\mathcal{U}_k$, $\forall\,k\in\mathbb{N}$; functions $G^{i}_{z,k}(z)\in\mathbb{R}$ for $z=x,u$, are assumed to be $C^2$; $f$ is a smooth state transition function and $f(0,0)=0$.
	
	In principle, different types of state constraints can be formalized as follows. For instance, (i) $\mathcal{X}_k$ with $G^{i}_{x,k}(x)=E_{k}^ix-c^i_k$ is a linear convex set,  where $E_{k}^i\in\mathbb{R}^{1\times n}$ and $c^i_k\in\mathbb{R}$ are time-varying parameters; (ii)
	$\mathcal{X}_k$ with $G^{i}_{x,k}(x)=d_k^i-\|E_k^ix_k-c^i_k\|$ represents a dynamic obstacle avoidance constraint of a robot in a 2-D map, where $E_k^i\in\mathbb{R}^{2\times n_i}$,  $c_k^i\in\mathbb{R}^2$ and $d_k^i\in\mathbb{R}$ are the center and radius of the circular dynamic obstacle respectively.
	
	\begin{definition}[Local stabilizability~\cite{zhang2021robust}]\label{Eqn:cost-finite}
		System~\eqref{Eqn:LL} is stabilizable on  $\mathcal{X}_k\times \mathcal{U}_k$ if, for any $x_0\in\mathcal{X}_k$, there exists a $C^1$ state-feedback policy $u(x_{k})\in\mathcal{U}_k$, $u(0)=0$, such that $x_{k}\in\mathcal{X}_k$ and $x_{k}\rightarrow 0$ as $k\rightarrow+\infty$. 
	\end{definition}
	{\color{black}
		\begin{assumption}[Lipschitz continuous]\label{ASSM:LIPSCHITZ}
			Model~\eqref{Eqn:LL} is Lipschitz continuous in $\mathcal{X}_k\times \mathcal{U}_k$, for all $k\in\mathbb{N}_1^{\infty}$, i.e., there exists a Lipschitz constant $0<L_f<+\infty$ such that for all $ x_1,x_2\in \mathcal{X}_k$ and $C^1$ control policies with $u(x_1),u(x_2)\in \mathcal{U}_k$,
			\begin{equation}\label{Eqn:lipschitz}
			\|f(x_1,u(x_1))-f(x_2,u(x_2))\|\leq L_f\|x_1-x_2\|.
			\end{equation}
	\end{assumption}}
	\begin{assumption}[Model]\label{assum:f}
		$\|\triangledown_u f(x,u)\|\leq g_m$ in the domain $\mathcal{X}_k\times \mathcal{U}_k$, where $g_m$ {\color{black}is a positive scalar}. 
	\end{assumption}
	\subsection{Control Objective}
	Starting from any initial condition $x_0\in\mathcal{X}_0$, the control objective is to find an optimal control policy $u^{\ast}_{k}=u^{\ast}(x_{k})\in\mathcal{U}_k$ that minimizes a quadratic regulation cost function of type
	\begin{equation}\label{Eqn:cost}
	J(x_0,u_{0:+\infty})=\sum_{k=0}^{+\infty} \gamma^{k}r(x_{k},u_{k}),
	\end{equation}
	subject to model~\eqref{Eqn:LL}, $x_k\in\mathcal{X}_k$, and $u_k\in\mathcal{U}_k$, $\forall k\in\mathbb{N}$;
	where 
	\begin{equation}\label{Eqn:stagecost}r(x_{k},u_{k})=\|x_{k}\|_{Q}^2+\|u_{k}\|_R^2,
	\end{equation}
	and $Q=Q^{\top}\in\mathbb{R}^{n\times n}$, $R=R^{\top}\in\mathbb{R}^{m\times m}$, $Q,R\succ 0$, $\gamma$ is a discounting factor.
	
	Without loss of generality, many waypoint tracking problems in the robot control field can be naturally formed as the prescribed regulation one, with a proper coordination transformation of the reference waypoints. 
	More generally, it is allowed that the time-varying state constraint might not contain the origin for some $k\in\mathbb{N}$. Typical examples can be found in, for instance, path following of mobile robots with collision avoidance, where the potential obstacle to be avoided might occupy the reference waypoints, i.e., the origin after coordination transformation.  In this scenario, it is still reasonable to introduce the following assumption for convergence guarantee.
	\begin{assumption}[State constraint]\label{assum:state-con}
		There exists a finite number $\bar k\in\mathbb{N}$ such that  $\{0\}\subseteq\mathcal{X}_k$  as $k\geq\bar k$. 
	\end{assumption}
	\begin{definition}[Multi-step safe control]
		For a given state $x_{k}\in\mathcal{X}_k$ at time instant $k$, a control policy $u(x_{k})\in\mathcal{U}_k$, is $L$-step safe for~\eqref{Eqn:LL} if the resulting future state evolutions of~\eqref{Eqn:LL} under $u(x_{k})$ satisfy $x_{k+i}\in\mathcal{X}_{k+i}^u$, $\forall i\in\mathbb{N}_1^L$, where $\mathcal{X}_{k+i}^u$ is the resulting state constraint under $u(x_k)$.   	
	\end{definition}
	
	To simplify the notation, in the rest of the paper, the super index in $\mathcal{X}_{k}^u$ is neglected, i.e., we use $\mathcal{X}_{k}$ to denote $\mathcal{X}_{k}^u$.

	\subsection{Preliminary Reinforcement Learning Solutions}
	In a special case where only control constraint is considered, i.e., $\mathcal{X}_k=\mathbb{R}^n$,  the optimal value function can be defined by
	$$J^{\ast}(x_{k})=\min{u_{k}\in\mathcal{U}_k,\forall k\in\mathbb{N}}\sum_{k=0}^{+\infty} \gamma^{k}r(x_{k},u(x_k)),$$
	which satisfies the Hamilton-Jacobi-Bellman (HJB) equation
	\begin{equation}\label{Eqn:cost-boot-optimal}
	J^{\ast}(x_{k})=\min{u_k\in\mathcal{U}_k} r(x_{k},u(x_{k}))+\gamma J^{\ast}(x_{k+1}),
	\end{equation}
	and the optimal control policy is
	\begin{equation}\label{Eqn:control-optimal}
	\begin{array}{ll}
	u^{\ast}(x_k)=\argmin{u_k\in\mathcal{U}_k} r(x_{k},u(x_{k}))+\gamma J^{\ast}(x_{k+1}).
	\end{array}
	\end{equation}
	Various RL solutions have been contributed to solving optimization problem~\eqref{Eqn:cost-boot-optimal} with~\eqref{Eqn:control-optimal}, resorting to an actor-critic approximating structure (cf.~\cite{zhang2009neural,liu2013policy,luo2017output}). 
	In control problems with state constraints, solving~\eqref{Eqn:cost-boot-optimal} with~\eqref{Eqn:control-optimal} with safety guarantee is more challenging using the trial-and-error-based actor-critic reinforcement learning framework. Recently, various safe RL solutions have been emerged~\cite{chow2019lyapunov,berkenkamp2018safe,garcia2015comprehensive,srinivasan2020learning,yu2019convergent,xu2021crpo,ma2021model,yang2021accelerating,huh2020safe,zheng2021safe,marvi2021safe,chen2021context,li2021safe,brunke2021safe,richards2018lyapunov}. 
	However, few works have addressed the safe RL algorithm design under time-varying safety constraints. 
	
	{\color{black}One of the typical safe RL algorithms is to shape the cost~\eqref{Eqn:cost} with barrier functions (see~\cite{geibel2005risk,balakrishnan2019structured,hu2020learning,tessler2018reward}), 
		which could be insufficient in many cases to guide how to ensure safety by actor-critic learning. Take the considered discrete-time optimal control problem as an example. Note that the gradients of barrier functions change rapidly when the state or control approach constraints boundaries.  On this ground, the reshaped cost function with barrier functions might experience abrupt changes since it is evaluated at discrete-time instants (in most cases the sampling interval is not chosen small enough). As a result, the weights of the actor-critic networks are prone to divergence due to the abrupt cost value changes in the training process.} 
	For these reasons,  we propose a safe RL approach using a BCP structure and an MPE mechanism for optimal control under time-varying safety constraints. 
	\subsection{Definition on Barrier Functions}
As policy improvement is usually performed by the gradient descent method in actor-critic RL, we have to reconstruct the cost function in~\eqref{Eqn:cost} by incorporating continuous barrier functions of state and control constraints.  To this end, we first introduce a definition of barrier functions as follows. 
	\begin{definition}[Barrier function~\cite{wills2004barrier}]\label{defi:polytopic}
		For a general convex set $\mathcal{Z}_k=\{z\in\mathbb{R}^l|G^i_{z,k}(z)\leq0,\,\forall i\in\mathbb{N}_1^{p_z}\}$, a barrier function is defined as
		\begin{equation}\label{Eqn:barrier_in}
		{\mathcal{B}}^{o}_k(z)=\left\{\begin{array}{l}
		-\sum_{i=1}^{p_z}{\rm{log}}\big(- G^{i}_{z,k}(z)\big),\ \ z\in {\rm {Int}}(\mathcal{Z}_k)\\
		[0.2cm]+\infty\ \ \rm{otherwise}.
		\end{array}\right. \qquad
		\end{equation}
	\end{definition} 
	
	{\color{black} To derive a satisfactory control performance, we define a recentered transformation of $ {\mathcal{B}}_k^o(z)$ centered at $z_c\in\mathbb{R}^l$ is defined as 
		${\mathcal{B}}_k^c(z)= {\mathcal{B}}^o_k(z)- {\mathcal{B}}_k^o(z_c)-\triangledown_z{\mathcal{B}}^o_k(z_c)^{\top}z$,  where $z_c=0$ if $\{0\}\subseteq \mathcal{Z}_k$ or  $z_c$ is selected such that $z_c\in\mathcal{Z}_k$ otherwise. }
	This definition leads to  the property that ${\mathcal{B}}^c_k(z)\geq 0$ and it reaches the minimum at $z_c$, i.e.,
	\begin{equation}\label{Eqn:origin-barrier}
	{\mathcal{B}}^c_k(z_c)=0,\ \triangledown {\mathcal{B}}^c_k(z_c)=0.
	\end{equation}
	For the case $\{0\}\nsubseteq\mathcal{Z}_k$, we suggest selecting $z_c$ far from the set boundary of $\mathcal{Z}_k$ and as the central point or its neighbor of $\mathcal{Z}_k$ (if possible).
	\begin{lemma}[Relaxed barrier function~\cite{wills2004barrier}] \label{lemma:relax}\hfill
		Define a relaxed barrier function of ${\mathcal{B}}_k^c(z)$ as
		\begin{equation}\label{Eqn:relaxed_B}
		{\mathcal{B}}_k(z)=\left\{\begin{array}{ll}
		{\mathcal{B}}_k^c(z)&\bar\sigma_k\geq \kappa_b\\
		\gamma_{b}(z,\bar\sigma_k)&\bar\sigma_k <\kappa_b,
		\end{array}\right.
		\end{equation}
		where the relaxing factor $\kappa_b>0$ is a small positive number, 
		$\bar\sigma_k={\rm min}_{i\in\mathbb{N}_{1}^{p_z}}-G^{i}_{k}(z)$, the function $\gamma_{b}(z,\bar\sigma_k)$ is strictly monotone and differentiable on $(-\infty,\kappa_b)$, and $\triangledown_z^2\gamma_{b}(z,\bar\sigma_k)\leq \triangledown_z^2{\mathcal{B}}_k(z)|_{\bar\sigma_k= \kappa_b}$,  then there exists a  matrix $H_{z_k}\geq\triangledown_z^2{\mathcal{B}}_k(z)|_{\bar\sigma_k= \kappa_b},$ such that $\|\triangledown_z {\mathcal{B}}_k(z)\|\leq {\mathcal{B}}_{z_k,m},$  ${\mathcal{B}}_{z_k,m}=\max_{z\in\mathcal{Z}_k}\|2H_{z_k}(z-z_c)\|$. 
		%
	\end{lemma}
	\textbf{Proof}: For details please see~\cite{wills2004barrier}. 
	\hfill $\square$
	
	\section{Safe RL with BCP and MPE}
	This section presents our safe RL approach and its implementation by an efficient barrier-based actor-critic learning algorithm.  Our safe RL approach has two novel ingredients. The first is a barrier force-based control policy structure, which has  physics force interpretations to ensure safety. The second is a multi-step policy evaluation mechanism, which provides the multi-step safety risk prediction to guide the policy to update safely online under time-varying constraints.
	\subsection{Design of Safe RL with BCP and MPE}\label{sec:31}
	We reconstruct the performance index $J(x_{k})$ with state and control barrier functions defined in~\eqref{Eqn:barrier_in}. Letting $\mu>0$ be a tuning parameter, the resulting value function, denoted as $\bar J(x_{k})$, is defined as
	\begin{equation}\label{Eqn:re-cost}
	\begin{array}{cl}
	\bar J(x_{k})=\sum_{k=0}^{+\infty} \gamma^{k}\bar r(x_{k},u_{k}),
	\end{array}
	\end{equation}	
	where $\bar r(x_{k},u_{k})=r(x_{k},u_{k})+\mu\mathcal{B}_k(u_{k})+\mu\mathcal{B}_k(x_{k})$. Note that, in
	addition to the logarithmic barrier function~\eqref{Eqn:barrier_in}, other general
	types of differentiable barrier functions such as exponential,
	polynomial ones can be naturally used instead to construct $\bar J(x_{k})$; which is, however,  beyond the scope of this work. 
	
	%
	{\color{black}
		\begin{proposition}[Unconstrained control problem equivalence]\label{prop:1}
			Letting $s_k(x_k)=\sqrt{ \mathcal{B}_k(x_{k})}$, the control problem for~\eqref{Eqn:LL} with cost~\eqref{Eqn:re-cost} is equivalent to an unconstrained optimal control problem for the system
			\begin{equation}\label{Eqn:extend-state}
			\left\{\begin{array}{ll}
			x_{k+1}=f(x_k,u_k)\\
			y_k=(x_k,s_k(x_k)),
			\end{array}\right.
			\end{equation}
			with the cost in~\eqref{Eqn:re-cost} being rewritten as
			\begin{equation}\label{Eqn:extend-cost}
			\bar J_u(x_{k})=\sum_{k=0}^{+\infty} \gamma^{k}(\|y_k\|_{Q_y}^2+\|u_k\|_R^2+\mu\mathcal{B}_k(u_{k})),
			\end{equation}
			where $Q_y={\rm{diag}}\{Q,\mu\}$. 
		\end{proposition}
		\textbf{Proof}. First, note that $\bar J_u$ in~\eqref{Eqn:extend-cost} is equivalent to $\bar{J}$ in~\eqref{Eqn:re-cost} since $y_k=(x_k,s_k(x_k))$ and $s_k(x_k)=\sqrt{ \mathcal{B}_k(x_{k})}$. Hence,  the control problem for~\eqref{Eqn:extend-state} with cost~\eqref{Eqn:extend-cost} is equivalent to that for~\eqref{Eqn:extend-state} with cost~\eqref{Eqn:re-cost}, and consequently to that for~\eqref{Eqn:LL} with cost~\eqref{Eqn:re-cost} by disregarding the definition $y_k=(x_k,s_k(x_k))$ in~\eqref{Eqn:extend-state}.	
		\hfill$\square$}
	
	In~\eqref{Eqn:extend-cost},  the overall objective function consists of the classical quadratic-type regulation costs and the barrier functions on the state and control. 
	{\color{black}The tuning parameter $\mu$ determines the influence of the barrier function values on the overall objective function.  
		Given $\mu$, the barrier functions even become dominant if the control and state are close to the boundaries of the safety constraints. 	Indeed, the parameter $\mu$ (concerning $Q$ and $R$) represents a trade-off between optimality and learning safety.}
	
To solve the control problem with  $\bar J(x_{k})$, we propose a novel barrier force-based control policy inspired by the barrier method in interior-point optimization~\cite{boyd2004convex}, i.e.,
	\begin{equation}\label{Eqn:control}
	u_{k}=v_{k}+\rho\triangledown_v\mathcal{B}_k(v_{k})+K\triangledown_x\mathcal{B}_k(x_{k}),
	\end{equation}
	where  $v_k\in\mathbb{R}^m$ is a new virtual control input, $\rho\in\mathbb{R}$ and $K\in\mathbb{R}^{m\times n}$ are decision variables to be further optimized (see also Section~\ref{sec:32}); 
	$\triangledown_v\mathcal{B}_k(v_{k})$ is the gradient of
	$\mathcal{B}_k(v_{k})$ for $v_k\in\mathcal{U}_k$, 
	$\triangledown_x\mathcal{B}_k(x_{k})$ is the gradient of $\mathcal{B}_k(x_{k})$ for $x_k\in\mathcal{X}_k$.
	\begin{remark}
 In~\eqref{Eqn:control}, the roles of the second and third terms are to generate the repulsive forces, respectively, as the variables $x$ and $v$ move toward the corresponding boundary of the constraints.
	As a result, ~\eqref{Eqn:control} generates joint forces to exactly balance the forces associated with $J(x_{k})$ and the barrier functions in $\bar J(x_{k})$.  Hence, our control policy has physics force interpretations to ensure safety. 
		\end{remark}
		\begin{remark}\label{remark:compare}
			Different from cost shaping-based RL algorithms~\cite{geibel2005risk,yang2021wcsac,balakrishnan2019structured,hu2020learning,chow2019lyapunov,tessler2018reward} and primal-dual approaches~\cite{chen2016stochastic,paternain2019safe,xu2021crpo,ding2021provably}, in the proposed approach, the decision variables $v(x_k),\,\rho$, and $K$, instead of $u(x_k)$, are optimized (deferred in~\eqref{Eqn:policy-im-hdP}). Then the improved decision variables are used to compute the three control terms in~\eqref{Eqn:control} to construct the real control policy $u(x_k)$.
		\end{remark}

	Let at any time instant $k$, the $L$-step ahead control policy be $u(x_\tau)\,\forall \tau\in\mathbb{N}_0^{L-1}$ where $L\in\mathbb{N}$.  Hence, one can write the following  difference equation for the multi-step prediction of the stage cost under $u(x)$, i.e., 
	\begin{equation}\label{Eqn:V_bzk}
	\begin{array}{ll} \bar J\big(x_{k}\big)
	&=\bar r( x_{k},u( x_k))+ \gamma\bar J\big( x_{k+1}\big)\\
	&=\sum_{l=0}^{L-1} \gamma^l\bar r( x_{k+l},u( x_{k+l}))+ \gamma^L\bar J\big( x_{k+L}\big).
	\end{array}
	\end{equation}
	Under control~\eqref{Eqn:control}, letting $\bar J^{*}(x_{k})$ be the optimal value function at time instant $k$,  a variant of the discrete-time HJB equation can be written as
	\begin{equation*}\label{Eqn:HJB}
	\hspace{-2mm}
	\begin{array}{ll}
	\bar J^{*}(x_{k})\hspace{-2mm}&=\min{ u_k\in\mathcal{U}_k} \bar r(x_{k},u(x_k))+ \gamma\bar J^{\ast}\big(x_{k+1}\big),\\
	\hspace{-2mm}&=\hspace{-8mm}\min{ \begin{array}{c} \scriptscriptstyle u_{k+l}\in\mathcal{U}_{k+l},
		\scriptscriptstyle l\in\mathbb{N}_0^{L-1}
		\end{array}} \hspace{-4mm}\sum_{l=0}^{L-1}\gamma^l\bar r(x_{k+l},u(x_{k+l}))+ \gamma^L\bar J^{\ast}\big(x_{k+L}\big)
	\end{array}
	\end{equation*}
	and the optimal solution is
	\begin{equation*}\label{Eqn:hjb}
	\hspace{-2mm}
	\begin{array}{ll}
	u^{\ast}(x_k)\hspace{-2mm}&=\hspace{0mm}\argmin{ u_{k}\in\mathcal{U}_k} \bar r(x_{k},u(x_k))+ \gamma \bar J^{\ast}\big(x_{k+1}\big)\\
	\hspace{-2mm}&=\hspace{-7mm}\argmin{\begin{array}{c}\scriptscriptstyle u_{k+l}\in\mathcal{U}_{k+l}, \scriptscriptstyle l\in\mathbb{N}_0^{L-1}\end{array}} \hspace{-4mm}\sum_{l=0}^{L-1}\gamma^l\bar r(x_{k+l},u(x_{k+l}))+ \gamma^L\bar J^{\ast}\big(x_{k+L}\big).
	\end{array}
	\end{equation*}
	We propose a safe RL algorithm with barrier-based control policy (BCP) and multi-step policy evaluation (MPE) in Algorithm~\ref{alg:safe-hdp} to solve  $u^{\ast}(x_k)$ and $\bar J^{\ast}(x_k)$.
	\begin{algorithm}[h]
		\caption{Safe RL with BCP and MPE}
		\label{alg:safe-hdp}
		\begin{algorithmic}
			\REQUIRE $\bar{\epsilon}>0$, policy $u^0(x_k)$,  $i=0$.
			\FOR {$k=1,2,\cdots$}
			\WHILE{$\bar J^{i}_k-\bar J^{i-1}_k\geq \bar{\epsilon}$} 
			\STATE \textbf{1)} Compute $x_{k+l}$ with $u^i(x_{k+l})$ based on model~\eqref{Eqn:LL} for $l\in\mathbb{N}_1^L$.%
			\STATE \textbf{2)}  multi-step policy evaluation:
			\begin{subequations}\label{Eqn:hdp}
				\hspace{-2mm}
				\begin{align}\label{Eqn:value-up-hdp}
				\hspace{-6mm}\begin{array}{ll}
				\bar J^{i+1}\big(x_{k}\big)=\sum_{l=0}^{L-1}\gamma^l\bar r(x_{k+l},u^i(x_{k+l}))+ \gamma^L\bar J^i\big(x_{k+L}\big).
				\end{array}
				\end{align}	\hspace{-4mm}
				\STATE \textbf{3)} Barrier-based control policy update:
				\begin{align}\label{Eqn:policy-im-hdP}\hspace{-5mm}
				\begin{array}{ll}
				(v(x_k),\rho,K)^{i+1}=\argmin{v_k,\rho,K}\bar r(x_k,u(x_k))+ \gamma\bar J^{i+1}\big(x_{k+1}\big),\vspace{1mm}\\
				u^{i+1}(x_k)=v^{i+1}(x_k)+\rho^{i+1}\triangledown_v\mathcal{B}_k(v^{i+1}(x_k))\vspace{1mm}\\
				\hspace{55mm}+K^{i+1}\triangledown_x\mathcal{B}_k(x_{k}).
				\end{array}	
				\end{align}
			\end{subequations}
			\STATE \textbf{4)} $i\leftarrow i+1$.
			\ENDWHILE
			\ENDFOR%
		\end{algorithmic}
	\end{algorithm}
	\begin{remark}
		{\color{black}
			Note that model-free RL algorithms have received considerable attention in continuous control tasks~\cite{haarnoja2018soft}. However, model-free approaches still have the data-inefficient issue, suitable for specific tasks with valid datasets~\cite{ma2021model,cowen2020samba}. In our case, we focus on a model-based framework with safety and convergence guarantees because it is more suitable for our concerned real-world safety-critical vehicle control tasks. The extension of our approach to the model-free case will be left for further investigation.} 
	\end{remark}
	
	\subsection{Barrier-based Actor-Critic Implementation}\label{sec:actor-critic}
	In the following, Algorithm~\ref{alg:safe-hdp} is implemented with a barrier-based actor-critic learning structure. 
	We first construct a consistent type of critic network to $\bar J$ in~\eqref{Eqn:re-cost} with barrier functions:
	\begin{equation}\label{eqn:critic}
	\hat{\bar J}(x_{k})=W_{c1}^{\top}\sigma_{c}(x_{k})+W_{c2}\mathcal{B}_k(x_{k}), 
	\end{equation}
	where $W_{c1}\in\mathbb{R}^{N_c}$ and $W_{c2}\in\mathbb{R}$ are weighting matrices, 
	$\sigma_{c}\in\mathbb{R}^{N_{c}}$ is a vector composed of basis functions. In a collective form, we write $\hat{\bar J}(x_{k})=W_{c}^{\top}h_{c}(x_{k}),$ where $W_c=(W_{c1},W_{c2})
	$, $h_c(x_k)=(\sigma_{c}(x_k),\mathcal{B}_k(x_{k}))$. 
	
	The ultimate goal of the critic network is to minimize the distance between $\bar J^{\ast}$ and $\hat {\bar J}$ via updating $W_c$. However, as $\bar J^{\ast}$ is not available, the following $\bar J^d(x_k)$ (defined according to~\eqref{Eqn:value-up-hdp}) is used as the target to be steered by $\hat {\bar J}$, i.e.,
	\begin{equation}\label{Eqn:lam_d}
	\begin{array}{l}
	{\bar J}^d(x_{k})=
	\sum_{l=0}^{L-1}\gamma^l\bar r(x_{k+l},u_{k+l})+ \gamma^L\hat {\bar J}(x_{k+L}).
	\end{array}
	\end{equation}
	Let $ \varepsilon_{c,k}= {\bar J}^d(x_k)-\hat {\bar J}(x_k)$  be the approximation residual, $
	\delta_{c,k}=\varepsilon_{c,k}^2$, and  $\gamma_{c}$ be the learning rate, then
	the update rule of weight $W_{c}$ according to the gradient descent is given as
	\begin{equation}\label{Eqn:wc}
	W_{c,k+1}={W_{c,k}} - {\gamma_{c}}\frac{\partial\delta_{c,k}}{\partial W_{c,k}}. 
	\end{equation}
	
	We next design the actor network for learning the control policy~\eqref{Eqn:control} with the following form
	\begin{equation}\label{Eqn:actor}
	u(x_{k})=W_{a,\sigma}^{\top}\sigma_{a}(x_k)+ \hat K\triangledown_x\mathcal{B}_k(x_k))+\hat{\rho}\triangledown_v\mathcal{B}_k(v_k), 
	\end{equation}%
	where $W_{a,\sigma}\in\mathbb{R}^{N_u\times m}$, $\hat K\in\mathbb{R}^{m\times n}$, and $\hat{\rho}\in\mathbb{R}$ are the weighting matrices, $\sigma_{a}\in\mathbb{R}^{\scriptscriptstyle N_{u}}$ is a vector of basis functions.  Let $W_{a}^{\top}=[W_{a1}^{\top} \, \hat{\rho}I]$, $W_{a1}^{\top}=[W_{a,\sigma}^{\top}\ \hat K]$ and $h_{a}(x_k)=(h_{a1}(x_k),\triangledown_v\mathcal{B}_k(v_k))$, $ h_{a1}(x_k)= (\sigma_{a}(x_k), \triangledown_x\mathcal{B}_k(x_k))$, then one can write~\eqref{Eqn:actor} in a collective form as $u(x_{k})=W_{a}^{\top}h_a(x_k)$.
	
\begin{figure}[h]%
	\centering
	\includegraphics[width=0.45\textwidth]{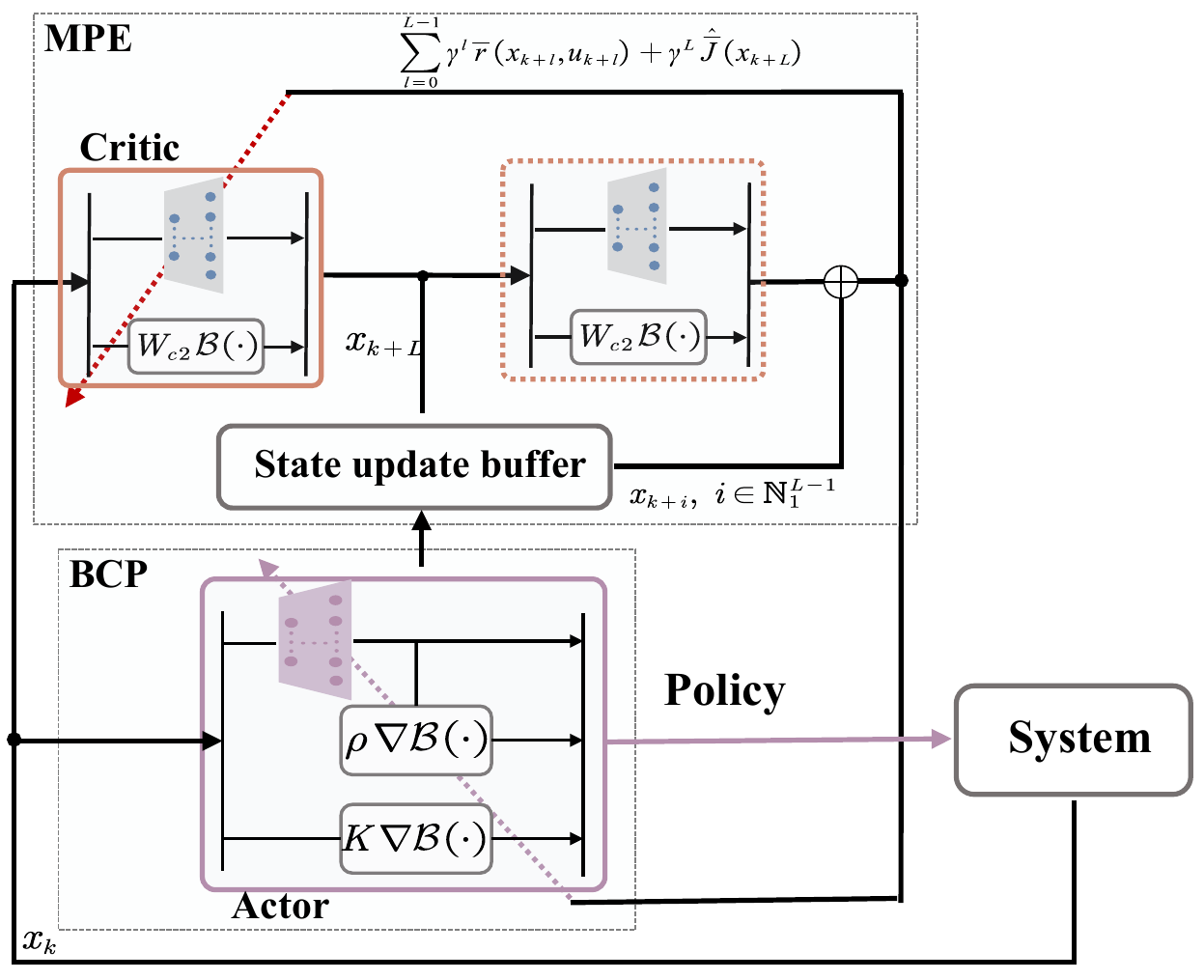}
	\caption{ The schematic diagram of the barrier-based actor-critic learning algorithm.
	}
	\label{fig:control-dia}
\end{figure}

	In view of~\eqref{Eqn:policy-im-hdP} and~\eqref{Eqn:actor},  letting $\nu_k=2Ru_k+\mu\triangledown_u \mathcal{B}_k(u_k)$, we define  a desired target of $\nu_k$, i.e., $\nu^d_k$ as
	$\nu_k^d
	=-\triangledown_u f(x,u)^{\top}\partial \hat {\bar J}(x_{k+1})/\partial x_{k+1}.$ 
	Denote $\varepsilon_{a,k}=\nu^d_k-\nu_k$ as the approximation residual,
	$\delta_{a,k}=\|\varepsilon_{a,k}\|^2$, and $\gamma_{a}$ be the learning rate, then
	the update rule of $W_{a1}$ and $\hat \rho$ according to the gradient descent is given as
	
	\begin{subequations}\label{Eqn:wa}
		\begin{align}
		W_{a1,k+1}=&{W_{a1,k}} - {\gamma_{a}}\frac{\partial\delta_{a,k}}{\partial W_{a1,k}},\\
		\hat{\rho}_{k+1}=&{\hat{\rho}_{k}} - {\gamma_{a}}\frac{\partial\delta_{a,k}}{\partial \hat{\rho}_{k}}.
		\end{align}
	\end{subequations}
For a visual display of the barrier-based actor-critic learning algorithm, please see Fig.~\ref{fig:control-dia}. 
	\begin{remark} 
		{\color{black}Although we introduce barrier-based terms in the actor-critic structure for policy learning, the resulting algorithm implementation procedure is comparable (only slightly more complex) to a standard actor-critic structure since all the weights can be learned with standard gradient decent rules (see~\eqref{Eqn:wc} and~\eqref{Eqn:wa}) and only a few more weighting matrices $W_{c,2}$, $\hat K$, and $\hat \rho$ are to be updated alongside.}    
	\end{remark}
	%
	\section{Theoretical results}\label{sec:32}
	The theoretical properties of the proposed safe RL in nominal and disturbance scenarios are proven in Section~\ref{sec:safety} and Section~\ref{sec:robustness} respectively. Then, the convergence analysis of the barrier-based actor-critic learning algorithm under time-invariant constraints is given in Section~\ref{sec:bac}.
	
	\subsection{Safety and Stability Guarantees in Nominal Scenario}\label{sec:safety}
In the following,	we prove the convergence of our proposed safe RL with BCP and MPE, i.e., Algorithm~\ref{alg:safe-hdp} under time-varying constraints. 
	\begin{assumption}[Stabilizability]\label{assum:stabilizable}
		For any $x_k\in\mathcal{X}_k$, there exist  $v(x_k)$, $\rho$, $K$ constituting a control policy $u(x_k)\in\mathcal{U}_k$ such that system~\eqref{Eqn:extend-state}  is locally stabilizable under~\eqref{Eqn:control}. 
	\end{assumption}

	Specifically, by setting $\rho=0$ and $K=0$, Assumption~\ref{assum:stabilizable} is equivalent to the standard local stabilizability condition as in~\cite{zhang2021robust} when the state and control constraints are time-invariant.  From Assumption~\ref{assum:stabilizable},  one can promptly derive the following $L$-step safe control condition: given $x_k\in\mathcal{X}_k$, there exists an $L$-step safe control policy such that $x_{k+l}\in\mathcal{X}_{k+l}$ $\forall l\in\mathbb{N}_1^L$. Note that this condition is equivalent to the existence of a 1-step safe control policy since the former can be derived from the latter using mathematical induction.
	To verify the above safe control condition, the variation of the state constraints can not be arbitrarily large. Let $\tilde{\mathcal{X}}_{k+1}=\{x_{k+1}|x_{k+1}=f(x_k,u_k),\, \forall x_k\in\mathcal{X}_k,u_k\in\mathcal{U}_k\}$ be the maximal reachable set from $\mathcal{X}_k$ under $\mathcal{U}_k$. We require that the state constraint at the any time $k+1$ satisfy  $\mathcal{X}_{k+1}\subseteq\tilde{\mathcal{X}}_{k+1}$. 
	\begin{theorem}[Convergence]\label{Eqn:theo-con}
		If $u^0(x_k)\in\mathcal{U}_k$ is such that the relaxed barrier function $\mathcal{B}_{k+l}(x_{k+l})$ is finite\footnote[1]{\label{condition} This implies that we do not require the initial control policy being $L$-step. This is due to the relaxed barrier function design in Lemma~\ref{lemma:relax}, where a smoother quadratic term is introduced to replace the barrier function when the variable  approaches the set boundary or is out of the set.}  $\forall l\in\mathbb{N}_1^L$, and the value function $\bar J^0(x_k)\geq\bar r(x_{k},u^0(x_k))+\gamma\bar J^{0}(x_{k+1})$; then with~\eqref{Eqn:hdp}, it holds that
		\begin{enumerate}[(i)]
			\item $\bar J^{i+1}(x_k)\leq V^i(x_k)\leq \bar J^{i}(x_k)$, where $V^i(x_k)=\bar r(x_{k},u^i(x_k))+\gamma\bar J^{i}(x_{k+1})$;
			\item  $\bar J^i(x_k)\rightarrow \bar J^{\ast}(x_k)$ and  $u^i(x_k)\rightarrow u^{\ast}(x_k)$, as $i\rightarrow +\infty$. 
		\end{enumerate}
	\end{theorem}
	\textbf{Proof}. Please refer to Appendix~\ref{sec:theorem}.  \hfill $\square$

\begin{remark}\label{remark-j}
		Let $J^{p}(x_k)$ be a safe and optimal value of~\eqref{Eqn:V_bzk} with $\mu=0$ and~\eqref{Eqn:control} under the control and state constraints. Then $\bar J^{\ast}(x_k)$ is a good approximation of $J^{p}(x_k)$ given $\mu$ being small. Moreover, if the optimization problem with~\eqref{Eqn:cost} under~\eqref{Eqn:control} is dual feasible, then one can obtain a quantifiable condition $\bar J^{\ast}(x_k)- J^{p}(x_k)\leq o(\mu)$, where $o(\mu)$ is a function that decreases along with $\mu$ (see Page 566 in~\cite{boyd2004convex}). This implies that the control policy $ u^{\ast}(x_k)$, associated with $\bar J^{\ast}(x_k)$,  is  $L$-step safe provided $\mu$ being chosen suitably small. 
	\end{remark}
	\begin{remark}
		Theorem~\ref{Eqn:theo-con} also implies that, at any time instant  $k$, the initial control policy is not necessarily $L$-step safe to guarantee safety and convergence. Hence, the recursive feasibility under time-varying constraints is likely guaranteed as long as an $L$-step safe control policy exists provided any $x_k\in\mathcal{X}_k$, which can be certified by Assumption~\ref{assum:stabilizable}.
	\end{remark}
	\begin{proposition}[Stability]\label{prop:stable}
		Let $\gamma=1$, $x_0\in\mathcal{X}_0$, and  $u^{\ast}$ be the control policy with the optimal solution $v^{\ast}$, $\rho^{\ast}$, and $K^{\ast}$, solved via minimizing~\eqref{Eqn:extend-cost} with~\eqref{Eqn:extend-state}. Under Assumptions~\ref{assum:state-con} and~\ref{assum:stabilizable}, the state $x_k$ of model~\eqref{Eqn:LL} using $u^{\ast}$, converges to the origin as $k\rightarrow +\infty$.
	\end{proposition}
	\textbf{Proof}. Please refer to Appendix~\ref{sec:theorem}. \hfill$\square$
	\begin{remark}
		Note that the discounting factor $0<\gamma\leq1$ is a crucial ingredient in reinforcement learning to ensure the convergence of the value function and control policy to the optimal ones (see~\cite{rizvi2018output}). However, as illustrated by Proposition~\ref{prop:stable}, a choice of $\gamma$ close to 1 is suggested to guarantee closed-loop stability. 
	\end{remark}
	{\color{black}
		\subsection{Safety and Robustness Guarantees in Disturbed Scenario}\label{sec:robustness}
		We show that our approach can guarantee safety and robustness under disturbances by  properly shrinking the state constraints in the learning process. To this end, let the real model dynamics be given as 
		\begin{equation}\label{eqn:perturbedsys}
		z_{k+1}= f(z_{k},u_{k})+w_k,
		\end{equation}
		where $z_k$ is the real state, $w_k\in\mathcal{W}$ is an additive bounded and unknown disturbance which can represent the modeled uncertainty or measurement noise, $\mathcal{W}$ is a compact set containing origin in the interior.
		In case the model dynamics are unavailable, the derivation of the nominal model~\eqref{Eqn:LL} may resort to data-driven modeling approaches. For a specific data-driven modeling approach and the estimation of the associated uncertainty set $\mathcal{W}$ please refer to~\cite{zhang2021robust}. 
		
		Let at any time instant $k$, $x_{k+j|k}$ be the predicted state by applying the control $u(x_k),\cdots, u(x_{k+L-1})$ using model~\eqref{Eqn:LL}. Assuming that  the uncertainty set $\mathcal{W}$ is norm-bounded, i.e., $\|w_k\|\leq \varepsilon_w$, then the following lemma is stated.
		\begin{lemma}[\cite{marruedo2002input}]\label{Eqn:lemma2}
			The difference between the real state under $u(z)$ and the nominal one under $u(x)$ satisfies 
			\begin{equation}
			\|x_{k+j|k}-z_{k+j}\|\leq \frac{L_f^j-1}{L_f-1}\varepsilon_w,
			\end{equation}
			where $x_{k|k}=z_{k}$. 
		\end{lemma}
		\textbf{Proof}. The proof is similar to~\cite{marruedo2002input}.  \hfill $\square$

		Let the constraint on the nominal state be shrink, i.e.,  $x_{k+j|k}\in\bar{\mathcal{X}}_{k+j}$ where $\bar{\mathcal{X}}_{k+j}=\mathcal{X}_{k+j}\ominus \mathcal{D}_{\varepsilon_w}^j$, $\mathcal{D}_{\varepsilon_w}^j=\{y\in\mathbb{R}^n|\|y\|\leq \frac{L_f^j-1}{L_f-1}\varepsilon_w\}$. The barrier function on the state in~\eqref{Eqn:re-cost} is modified according to the constraint $x_{k+j|k}\in\bar{\mathcal{X}}_{k+j}$. Assume that  the computed $\bar{\mathcal{X}}_{k+j}$ is non-empty and contains the origin in the interior for all $k\geq\bar k$. 
		\begin{theorem}[Robustness]\label{theorem:2} Under Assumptions~\ref{assum:state-con}-\ref{assum:stabilizable},
			the state evolution of~\eqref{eqn:perturbedsys}, by applying the  learned optimal policy $u^{\ast}$ with~\eqref{Eqn:LL}, converges to the set  $\mathcal{D}_{\varepsilon_w}^{\infty}$, i.e., $\lim_{k\rightarrow +\infty} x_k\rightarrow \mathcal{D}_{\varepsilon_w}^{\infty}$. 
		\end{theorem}
		\textbf{Proof}. Please refer to Appendix~\ref{sec:theorem}.
		\hfill $\square$
		
		As suggested in~\cite{marruedo2002input}, to reduce the size of $\mathcal{D}_{\varepsilon_w}^j$, i.e., the Lipschitz constant $L_f$,   two design choices are suggested: (i) a different suitable norm type can be used; (ii) an additional feedback term $K(z_k-x_k)$ can be added in the control input to reduce the conservativeness of the multi-step prediction of~\eqref{Eqn:LL}, where $K\in\mathbb{R}^{m\times n}$ is a stabilizing gain matrix  of~\eqref{Eqn:LL}. 
	}
	
	\subsection{Convergence Analysis of BAC Learning Algorithm}\label{sec:bac}
	Note that, as shown in the Proposition~\ref{prop:1},  the control problem for~\eqref{Eqn:LL} with $\bar J(x_{k})$ is equivalent to an unconstrained problem for a time-varying model~\eqref{Eqn:extend-state}
	with the~\eqref{Eqn:extend-state}. The convergence analysis for the BAC learning algorithm in this scenario would be much involved by Lyapunov method since the optimal weights of the actor and critic are time-dependent due to $y_k=(x_k,\sqrt{\mathcal{B}_k(x_{k})})$. For the sake of simplicity, we recall that a time-varying constraint can be partitioned into several segments of time-invariant ones. Hence, in the following,
	we prove the convergence of the BAC learning algorithm under time-invariant state and control constraints, i.e., $\mathcal{X}=\mathcal{X}_k$ and  $\mathcal{U}=\mathcal{U}_k$. That is, we prove that whenever the constraints are changed, our algorithm can eventually converge after some time steps. To this end, one first write
	$${\bar J}^{\ast}(x_{k})={W_{c}^{\ast}}^{\top}h_{c}(x_{k})+\kappa_c(x_k)$$
	$$u^{\ast}(x_k)={W_{a}^{\ast}}^{\top}h_{a}(x_{k})+\kappa_a(x_k),$$ where  $W_{c}^{\ast}$ and $W_{a}^{\ast}$ are constant weights, $\kappa_c$ and $\kappa_a$ are reconstruction errors.
	In view of the universal capability of neural networks with one hidden-layer, we introduce the following assumption on the actor and critic network.
	\begin{assumption}[Weights and reconstruction errors of BAC]\label{assum:network}\hfill
		\begin{enumerate}[1)]
			\item $\|W_{c}^{\ast}\|\leq W_{c,m}$, $\|\sigma_{c}(x)\|\leq \sigma_{c,m}$, $\|\triangledown_x\sigma_{c}(x)\|\leq \bar\sigma_{c,m}$, $\|\kappa_{c}(x)\|\leq  \kappa_{c,m}$;
			\item $\|W_{a}^{\ast}\|\leq W_{a,m}$, $\|\sigma_{a}(x)\|\leq \sigma_{a,m}$, $\|\kappa_{a}(x)\|\leq  \kappa_{a,m}$. \hfill $\blacktriangleleft$
		\end{enumerate} 
	\end{assumption}
	
	To state the following theorem in a compact form, we let $\tilde W_{\star}= W_{\star}^{\ast}- W_{\star}$, $\star=a,c$ in turns, denote $\Delta \bar h_{c,k}=\Delta h_{c,k}^{\top}\Delta h_{c,k}$, where $\Delta h_{c,k}=\gamma^Lh_{c,k+L}-h_{c,k}$, and use $q$ and $q^{+}$ to denote $q_k$ and $q_{k+1}$ respectively unless otherwise specified.  For simplicity, we assume  that $G^{i}(u)=E^iu$, $E^i\in\mathbb{R}^{1\times m}$.
	\begin{theorem}[Convergence of BAC learning]\label{THEOM:3}
		Under Assumptions~\ref{assum:f} and~\ref{assum:network}, if \begin{subequations}\label{Eqn:con-condition}
			\begin{align}\label{Eqn:actor-cond}
			R-\mu H_u\succ0\ {\rm and}\ I-3d_m(R+\mu H_u)^2\succ0,
			\end{align}
			where $d_m= 4\gamma_{a}(\sigma_{a,m}^2+\mathcal{B}_{v,m}^2+\mathcal{B}_{x,m}^2)$, and
			\begin{align}\label{Eqn:critic-cond}
			q_1\leq \Delta \bar h_{c,k}\leq q_2,
			\end{align}
			where $q_1,q_2>0$,
		\end{subequations}
		then it holds that
		$ \|(\xi_{a,k},\tilde W_{c,k})\|\leq \sqrt{\frac{\epsilon_{m}}{\lambda_{\rm min}(S)}},\ {\rm as}\, k\rightarrow+\infty,$
		where  $\xi_{a,k}=\tilde W_{a,k}^{\top}h_a(x_k)$, $\epsilon_{m}$  is a bounded error and $S$ is a positive-definite matrix, whose definitions are deferred in Appendix~\ref{sec:theorem}. Also,
		$(\xi_{a,k},\tilde W_{c,k})\rightarrow 0,\ {\rm as}\, k\rightarrow+\infty,$
		if $\kappa_{\star}(x_k)\rightarrow 0$, $\star=a,c$ in turns. 
	\end{theorem}
	\textbf{Proof}. Please refer to the Appendix. \hfill $\square$
	\section{Simulation and experimental results}
	The developed theoretical results are first verified with two robot simulated examples. Please see Appendix~\ref{sec:simulation} for the detailed implementation steps and results. 
	In this section, we focus on the applications of our approach to two real-world intelligent vehicles. {\color{black}Specifically, an integrated path following and collision avoidance problem is considered, which represents a crucial capability for navigation of intelligent vehicles under unknown and dynamic environments~\cite{sgorbissa2019integrated,lapierre2007combined}.}
	\begin{table*}[h]
		\centering \caption{Numerical comparisons  in Safety Gym with randomly generated obstacle positions.} 
		\label{tab:Tab_com-safetygym}
		
		\scalebox{0.7}{
			\begin{tabular}{@{}ccccccccc@{}}
				\toprule
				Approach&     Collision rate    & Target reach & Average speed (m/s)  &Training time (s) &Episode& Samples &Training scenario& Deployment scenario \\ 
				\midrule
				CPO
				&0.1 &0.9        & 0.82    & 2.4e4    &1000&3.3e5 & Safety Gym& Safety Gym   \\ 
				TRPO-L
				&0.095          & 0.905   &0.85  &2.8e4&1000&3.3e5 & Safety Gym& Safety Gym                 \\
				PPO-L
				&0.095    &0.905  &0.84       & 2.7e4   &1000&3.3e5 & Safety Gym& Safety Gym                 \\ 
				DDPG-CS
				&--    &--     &--   & 2.8e4  &1000&3.3e5  & with data from~\eqref{Eqn:discrete-time formation tracking error model}& --      \\  
				SAC-CS
				&--    &--    &--   & 2.8e4  &1000&3.3e5& with data from~\eqref{Eqn:discrete-time formation tracking error model}& --              \\ 
				{\multirow{2}{*}{Ours without MPE} }
				&\textbf{0.025}  &\textbf{0.975} & 0.76       & {1.7}  &\textbf{100}&\textbf{5e3}&with model~\eqref{Eqn:discrete-time formation tracking error model} &Safety Gym   \\   
				&\textbf{0.03}  &\textbf{0.97} & 0.8        & --  &--&--&--&-- \\                
				{Ours} 
				&\textbf{0}  &\textbf{0.945} & 0.76        & --  &--&--&--&-- \\                
				\bottomrule
		\end{tabular}}
	\end{table*}
	\begin{figure}[h]
		\centering
		\includegraphics[width=0.35\textwidth]{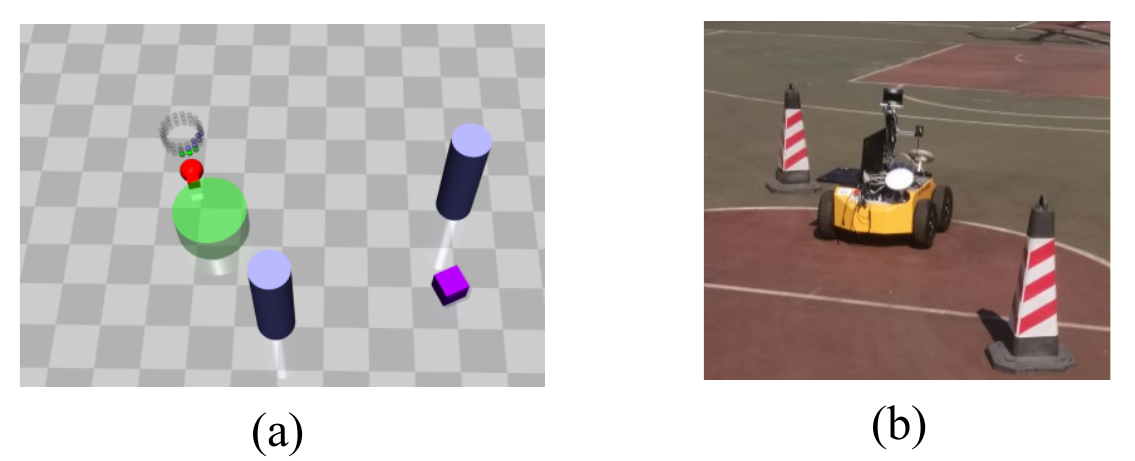}
		\caption{ (a) Simulation scenario in Safety Gym: the objective is to move the vehicle (red) to the green region while avoiding two static obstacles (grey), the moving soft object  (purple) is not considered in the controller design; (b) Experimental platform of the differential-drive vehicle and testing scenario.}
		\label{fig:safetygym}
	\end{figure}
	\begin{table*}[h]
		\centering \caption{Numerical comparisons in Safety Gym with generated obstacles on the path between the target and vehicle.
		}
		\label{tab:Tab_com-safetygym-2}
		
		\scalebox{0.7}{
			\begin{tabular}{@{}ccccccc@{}}
				\toprule
				Approach&     Collision rate  & Target reach   & Average speed (m/s)  &Training time (s) &Episode& Samples  \\ 
				\midrule
				CPO
				&0.815 &0.185 &0.76            & 2.4e4    &1000&3.3e5     \\ 
				TRPO-L
				&0.835 &0.165            & 0.8  &2.8e4&1000&3.3e5                 \\
				PPO-L
				&0.8 &0.2   &0.78         & 2.7e4   &1000&3.3e5                 \\ 
				Ours without MPE
				&\textbf{0.23}  &\textbf{0.77} & 0.76       & {1.7}  &\textbf{100}&\textbf{5e3}                 \\  
				Ours&\textbf{0}&\textbf{0.595}  &0.76        & --  &--&--                \\              
				\bottomrule
		\end{tabular}}%
	\end{table*}

	{\color{black}
		\subsection{Application to a Differential-Drive Vehicle: Offline Learning Scenario}\label{sec:app-differ}
		Consider the integrated path following and collision avoidance problem of a differential-drive vehicle.
		Its kinematics model is 
		\begin{equation}\label{Eqn:kinematic model}
		\dot{q}=(\dot{p}_{x},\dot{p}_{y},
		\dot{\theta}) =
		(v_o\cos \theta, 
		v_o\sin \theta,
		\omega),
		\end{equation}
		where $\text{(}p_{x},p_{y}\text{)}$ is the coordinate of the vehicle in Cartesian frame, $\theta$ is the yaw angle, $u=\left[ v_o,\omega \right] ^{\top}$ is the control input, where $v_o$ and $\omega$ are the linear velocity and yaw rate, respectively.
		
		Let us define the path following error as 
		$e=q_r -q$, where $q_r$ is the reference state.
		One can write the error model as
		\begin{equation}\label{Eqn:discrete-time formation tracking error model}
		\left\{ \begin{array}{l}
		\dot e_{x} =\omega e_{y} -v_o+ v_{o,r}\cos e_{\theta}\\
		\dot e_{y} =-\omega e_{x}+v_{o,r}\sin e_{\theta}\\
		\dot e_{\theta}=\omega_r -\omega,\,\,\\
		\end{array} \right.  
		\end{equation}
		where $(e_{x},e_{y},e_{\theta})=:e$,  $v_{o,r}$ and $\omega_r$ are the reference inputs. 
		
		In implementation, model~\eqref{Eqn:discrete-time formation tracking error model} was discretized with a sampling interval $\varDelta t=0.05s$  to derive the model like~\eqref{Eqn:LL}.
		The constraint for collision avoidance was typically formulated as $\mathcal{X}_k=\{(p_{x},p_{y})|\|(p_{x},p_{y})-c_k\|\geq d\}$, where $d$ and $c_k$ are the radius and center of the obstacle respectively. Also, the size of $\mathcal{X}_k$ was properly shrunk by increasing $d$  to account for uncertainties.
		In the training, the penalty matrices were selected as $Q=I$, $R=0.1$, $\mu=0.001$. The discounting factor $\gamma$ was $\gamma=0.95$. The relaxing factor $\kappa_b$ was $\kappa_b=0.05$. The basis functions $\sigma_c(x)$ and $\sigma_a(x)$ were chosen as hyperbolic tangent activation functions with $N_c=N_u=4$.  The step $L$ was chosen as $L=10$. Weights  $W_c$ and $W_a$  were initialized with uniformly random numbers. 
		
		\textbf{Simulation results using Safety Gym environment~\cite{ray2019benchmarking}.} We tested our approach in the Safety Gym environment with the MoJoCo simulator~\cite{todorov2012mujoco}. Our method was compared with some of the state-of-the-art safe RL algorithms: constrained policy optimization (CPO)~\cite{achiam2017constrained}, trust region policy optimization with Lagrangian methods (TRPO-L)~\cite{ray2019benchmarking}, proximal policy optimization with Lagrangian methods (PPO-L)~\cite{ray2019benchmarking}, deep deterministic policy gradient~\cite{lillicrap2015continuous} with cost shaping (DDPG-CS), and soft actor-critic (SAC)~\cite{haarnoja2018soft} with cost shaping (SAC-CS). Note that the safety-aware RL~\cite{yang2019safety} was not directly applicable in this case since it is nontrivial to find an invertible barrier function of the obstacle constraint. In the training stage, all the parameter settings of CPO, TRPO-L, and PPO-L were consistent with that in~\cite{ray2019benchmarking}. In DDPG-CS and SAC-CS, we used the same cost function as ours, i.e.,~\eqref{Eqn:re-cost}. We directly deployed the offline learned control policy using the kinematics model in implementation since we did not know the vehicle's dynamic model. All the comparative algorithms were trained and deployed using the same environment in Safety Gym. 
		The simulation results in Table~\ref{tab:Tab_com-safetygym} show that our approach outperforms all the comparative algorithms in data efficiency, collision avoidance, and performance
		(see the video details\footnote[2]{\label{web}\href{https://youtube.com/playlist?list=PLPE5-2sIdTlhk5r0VQr-66PBEqpAXvdAx}{https://youtube.com/playlist?list=PLPE5-2sIdTlhk5r0VQr-66PBEqpAXvdAx.} or \href{https://pan.baidu.com/s/1NxJ-zgD4ZdVvqIXQgJkCqg}{https://pan.baidu.com/s/1NxJ-zgD4ZdVvqIXQgJkCqg,} with extracting code: 9426.}). 
		\begin{remark}
			There are failures under our offline learned policy without MPE for the following reasons. First, the obstacles' locations were generated randomly without considering the vehicle's physical limit. Second, the estimated uncertainty is inaccurate since the inputs in Safety Gym are saturated values with unknown physical interpretations. These issues lead to Theorem~\ref{theorem:2} being not verified. In our case, one can guarantee safety (possibly with conservativeness) by activating our MPE mechanism even if the model is inaccurate, see Table~\ref{tab:Tab_com-safetygym} and~\ref{tab:Tab_com-safetygym-2}. 	
		\end{remark} 
		
		As shown in Table~\ref{tab:Tab_com-safetygym-2}, when the obstacles overlapped with the reference path between the target and vehicle, our approach offers a significant performance improvement compared with other adopted approaches\footref{web}.  The DDPG-CS and SAC-CS failed to obtain the converged policy after several training trials. Therefore we are unable to show the results in the Tables. In summary, our approach outperforms the comparative model-free safe RL approaches for the following two reasons. Firstly, the proposed barrier force-inspired control policy structure has a clear physical interpretation to guarantee safety online and improve the generalization ability. Secondly, our approach is model-based, facilitating multi-step policy evaluation online.

		\textbf{Real-world experimental results with comparisons to nonlinear MPC algorithms.}
		We also tested our proposed algorithm on a real-world differential-drive vehicle platform. The control task is to follow a predefined reference path (with $v_{o,r}=0.7$ m/s) while passing and avoiding collision with a moving object (vehicle) that is traveling along the reference path. In such a situation, the conflict between the goals of path following and collision avoidance leads to a challenging multi-objective control problem.
		
		In the experiment, the vehicle was equipped with a Laptop running Ubuntu in an Intel i7-8550U CPU@1.80 GHz. The sampling interval was set as $\varDelta t=0.1$s.  We directly deployed the offline learned policy of our approach to control the vehicle.  At each sampling instant, the onboard laptop computed the control input in real-time using the state information, which was periodically measured by the onboard satellite inertial guidance integrated positioning system (SIGIPS). To simplify the experimental setup, another wheeled vehicle following the reference path with a lower speed profile ($v_{o,r}=0.3$ m/s) was regarded as the obstacle to be avoided. Its position and velocity information was measured in real-time by SIGIPS and transmitted to the ego vehicle via a WIFI network.
		\begin{table}[h!tb]
			\centering \caption{{\color{black}Online average computational time in each time step (Unit=ms).}}
			\label{tab:Tab_com-time}
			\renewcommand\arraystretch{1.1}
			\scalebox{0.7}{
				\begin{tabular}{cccccc}
					\hline
					\multirow{2}{*}{Method}&	
					\multicolumn{2}{c}{Ours} &
					\multirow{2}{*}{NMPC-c}&\multirow{2}{*}{NMPC-e}&\multirow{2}{*}{NMPC-cbf}  \\
					\cline{2-3}
					& Online deploy & Online learning & \multicolumn{2}{c}{} \\
					\hline
					MATLAB&\textbf{0.3}& \textbf{2.2} & 1360 & 1360&1450\\	\hline
					C$^{++}$& \textbf{0.04}& \textbf{--} & 140 & 135&94 \\\hline
				\end{tabular}
			}%
		\end{table}
		{\color{black}The following MPC algorithms were adopted for comparison.
			\begin{enumerate}
				\item A nonlinear MPC algorithm with nonconvex circular constraints (NMPC-c), where the constraint was designed according to~\cite{JEWISON2015257}. The vehicle obstacle was approximated by a circle, i.e., we enforce constraint $(\Delta p_x)^2+(\Delta p_y)^2>d_o^2$ in NMPC-c, where $\Delta p_x$ and $\Delta p_y$ were deviations from the robot to the obstacles in the associated coordinate axes, $d_o=1$m.
				\item A nonlinear MPC algorithm with nonconvex ellipsoidal constraints according to~\cite{Bruno2019}. The vehicle obstacle was approximated by an ellipsoid, where the semi-major axis of the ellipsoid was in the direction of the reference path. The semi-major radius and semi-minor radius were computed as $1.517$m and $1.017$m respectively according to~\cite{Bruno2019}. 
				\item A nonlinear MPC algorithm with control barrier function~\cite{zeng9483029} (NMPC-cbf). The collision avoidance constraint is formulated by a control barrier function constraint, i.e., $h(k+1)-h(k)\geq -\eta h(k)$, where $h=(\Delta p_x)^2+(\Delta p_y)^2-d_o^2$ is a control barrier function, the parameter $\eta$ is a positive scalar which is properly tuned for fair comparisons (see Table~\ref{tab:Tab_com-suc0}).
			\end{enumerate}
			The stage costs of all the comparative MPC algorithms were designed the same as in~\eqref{Eqn:stagecost}, and the prediction horizon was set as $N_p=20$. According to~\cite{Bruno2019}, the following potential function   
			\begin{equation*}
			J_{\rm {p}}(k)=\sum_{j=0}^{N_p-1}\mu_p\frac{1}{(\Delta p_x(k+j))^2+(\Delta p_y(k+j))^2+\epsilon_p},
			\end{equation*}
			was additionally adopted to improve the collision avoidance performance in NMPC-c and NMPC-e,  $\epsilon_p$ was chosen as $0.0001$, and $\mu_p$ was tuned for fair comparisons (see Table~\ref{tab:Tab_com-suc0}).  
			All the MPC algorithms were solved at each sampling interval based on the CasADi toolbox~\cite{Andersson2019} with an \texttt{Ipopt} solver~\cite{wachter2006implementation}. All the algorithms were tested under different reference profiles.
			Experimental results under dynamic collision avoidance were illustrated in Table~\ref{tab:Tab_com-suc0} and  Figs.~\ref{fig:nmpcc}-\ref{fig:our} in Appendix~\ref{appen:experi-results} (see the video details\footnote[3]{\label{web1} \href{https://pan.baidu.com/s/1NxJ-zgD4ZdVvqIXQgJkCqg}{https://pan.baidu.com/s/1NxJ-zgD4ZdVvqIXQgJkCqg,} with extracting code: 9426.}). The results show that the NMPC-c and NMPC-e failed in realizing overtaking and followed behind the moving obstacle when the adopted reference points were dense, while our approach can realize conflict resolution in all scenarios. Also, our approach outperforms NMPC-c and NMPC-e in terms of the planning performance and path following performance (see Table~\ref{tab:Tab_com-suc0}). In addition to the unique policy design and learning mechanism of our approach, the performance improvement to the MPC algorithms is also due to the significant computational load reduction (see Table~\ref{tab:Tab_com-time}, and Fig.~\ref{fig:cputime} in Appendix~\ref{appen:experi-results}). To further show the effectiveness of our approach, we carried out extra tests by manually manipulating the moving obstacle to block the path of the ego vehicle when the latter reacted promptly to avoid collision successfully (see Fig.~\ref{fig:our-noncoor} in Appendix~\ref{appen:experi-results}).} 

			\begin{figure*}[h]%
			\centering
			\includegraphics[width=0.8\textwidth]{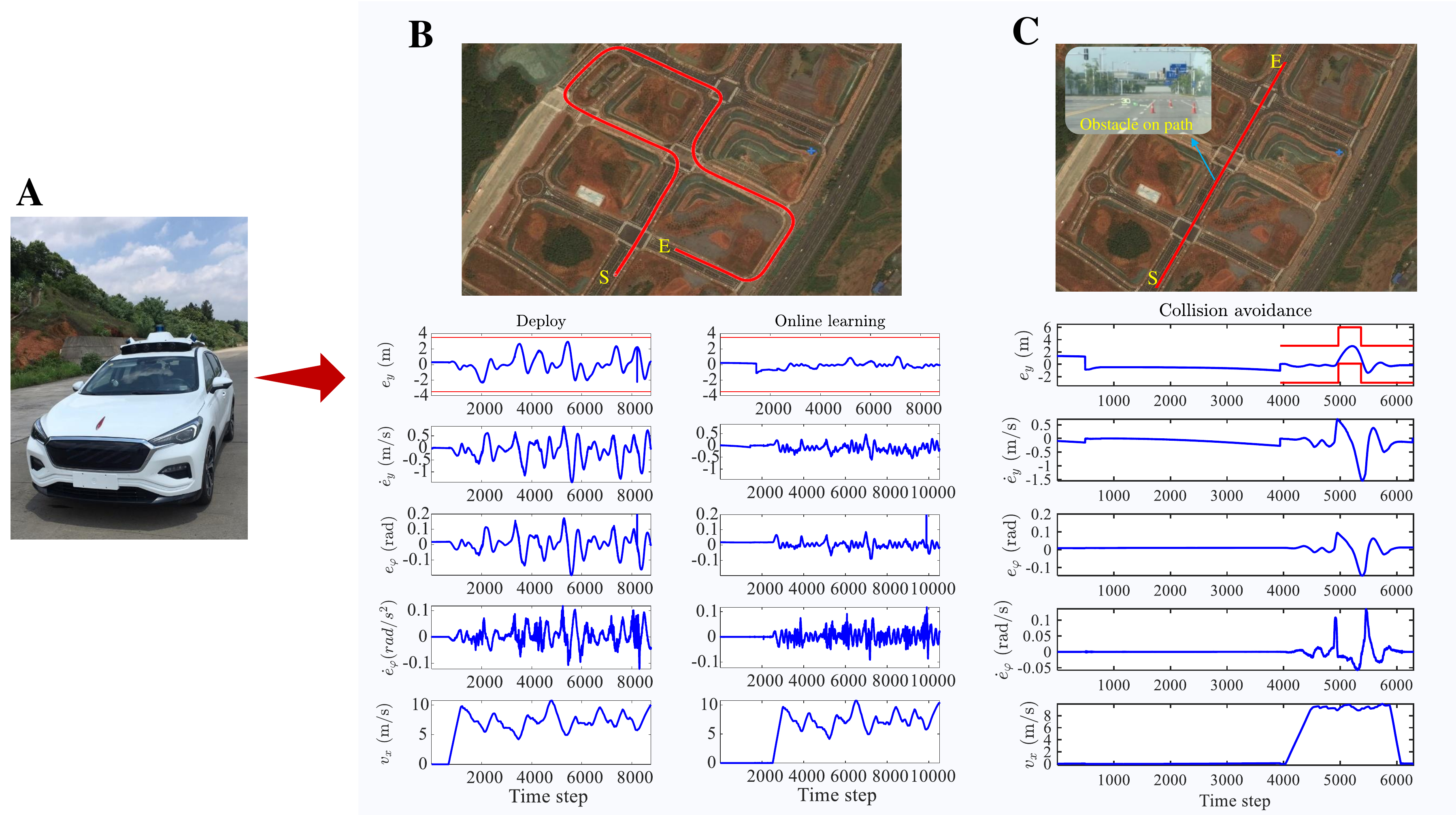}
			\caption{ A: The HongQi EHS3 autonomous driving experimental platform. B: The upper panel presents the road map with road boundary constraints, where ``S" stands for the starting point, ``E" stands for the ending point, and the red line is the reference path for path following control. The lower panel presents the corresponding state errors Compared with offline learning case, a significantly improved performance can be achieved by online policy learning. C: The upper panel presents the road map with collision avoidance scenario, while the lower panel gives the numerical state errors. 
			}
			\label{fig:experimental_platform}
		\end{figure*}
		\subsection{Application of an Ackermann-Drive Vehicle: Online Learning Scenario}
		Consider the path following control of an  Ackermann-drive vehicle with collision avoidance. Its simplified lateral dynamics is described by a  ``bicycle" model (cf.~\cite{rajamani2011vehicle}), that is
		\begin{align}\label{eq:stateDifferentialEquations}
		& \dot X = v_x \cos\varphi - v_y \sin\varphi \notag \\
		& \dot Y = v_x \sin\varphi + v_y \cos\varphi  \notag \\
		& \dot v_y = -v_x \dot\varphi + \frac{2}{m} \left[ C_{f} \left( \delta- \frac{v_y + l_f \dot\varphi}{v_x} \right) + C_{r} \frac{l_r \dot\varphi-v_y}{v_x} \right] \notag \\
		& \ddot\varphi = \frac{2}{I_z}\left[ l_f C_{f} \left( \delta - \frac{v_y + l_f \dot\varphi}{v_x} \right) - l_r   C_{r} \frac{l_r \dot\varphi-v_y}{v_x} \right], 
		\end{align}
		where $X$ and $Y$ are the coordinates of the vehicle center of mass in the Cartesian frame $XoY$, $v_x$ and $v_y$ are the longitudinal and lateral velocities respectively, $\varphi$ is the yaw angle, $I_z=4175$${\rm kg}\cdot m^2$ is the yaw moment of inertia, $m=1723$$\rm kg$ is the mass of the vehicle, $C_{f}=66900N$ and $C_{r}=62700 N$ are the cornering stiffness of the front and rear tires, respectively, $l_f=1.322m$, $l_r=1.468m$, $\delta$ is the front wheel angle variable to be manipulated. 

		Given path reference points $(X^r,Y^r)$ and $v_x$, we aim to minimize the lateral distance from the vehicle center of mass to the nearest reference point while  avoiding potential collisions with obstacles. To this end, let the nearest point be $(X_p^r,Y_p^r)$, then one can compute the reference yaw angle $\varphi^r_p$.  Define  $e_y=-(X-X_p^r){\rm sin}(\varphi_p^r)+(Y-Y_p^r){\rm cos}(\varphi_p^r),$ $e_{\varphi}=\varphi-\varphi_p^r$.
		Let $x=(e_y,\dot e_y,e_{\varphi},\dot e_{\varphi})$,  then one can obtain the continuous-time lateral dynamical model:  
		$\dot x=F_1(x)+F_2(x)\delta+F_3(x)\varphi_p^r,$
		where $F_1(0)=0$, $F_3(0)\neq 0$. Since $(x,\delta)=0$ might not be an equilibrium point if $\varphi_p^r\neq 0$, we introduced a virtual control variable $u=\delta+\delta_f$, where $\delta_f$ was selected such that $F_2(x)\delta_f=F_3(x)\varphi_p^r$. Consequently, the lateral dynamical model was discretized with a sampling interval $\Delta t=0.02s$, i.e.,
		$x_{k+1}=x_k+\Delta tF_1(x_k)+\Delta tF_2(x_k)u_k.$
		In the path following control task with collision avoidance,  the cost function was chosen as
		$\bar J=\sum_{k=0}^{+\infty}\|x_k\|_Q^2+\|u_k\|_R^2+\mu\mathcal{B}_k(e_{y,k}),$
		where $Q=I$, $R=1$, $\mu=0.02$. The basis functions $\sigma_c(x)$ and $\sigma_a(x)$ were chosen as polynomial kernel functions with $N_c=10$ and $N_u=14$.

		\textbf{Real-world experimental results\footref{web}.} We also tested our safe RL algorithm on the real-world intelligent vehicle platform built with a HongQi EHS3 electric car to realize the path following control (see Fig.~\ref{fig:experimental_platform}).  In the experiment, the states of the vehicle were measured by a SIGIPS; then, the measured states were transmitted to an industrial control computer, where the control policy was computed using our approach with a sampling interval of $0.02$s.  We first applied our algorithm to follow a reference path with road boundaries (see Fig.~\ref{fig:experimental_platform}-B). 
		Different from that in the simulation tests, the vehicle speed was controlled by a PI controller to track a time-varying speed reference. This caused a strong nonlinearity of the lateral dynamics, leading to extra difficulties in the control task. The experimental results displayed in Fig.~\ref{fig:experimental_platform} show that the control policy of our approach can be learned offline and deployed online safely, showing an impressive sim-to-real transfer capability. Also, one can achieve better control performance by online learning the control policy, which further demonstrates the adaptability of our approach to dynamic environments. 
		
		To show the capability of dealing with time-varying state constraints, we tested our approach to track a reference path that overlapped with obstacles (see Fig.~\ref{fig:experimental_platform}-C). Similarly, the location information of obstacles was assumed to be pre-detected. In the experiment, the control policy was learned and deployed synchronously online. The initial constraints were the road boundaries. Then, the constraint on $e_y$ was changed accordingly once the vehicle was near the obstacle. The vehicle using our approach can avoid collision successfully and converge rapidly to the reference path after completing the collision avoidance task (see again Fig.~\ref{fig:experimental_platform}).
		 
		{\color{black} 
			\subsection{Implementation Issues and Discussions}
			\textbf{Implementation issues}. First, the tuning parameter $\mu$  is suggested to be chosen smaller than the entries of $Q$ and $R$ to obtain a satisfactory control performance. A larger choice of $\mu$  might result in a safe but conservative control policy. Second, the initial values of $W_{a,\sigma}$, $\hat K$, and $\hat{\rho}$ in the actor must be properly selected such that the initial control policy with~\eqref{Eqn:actor} is $L$-step safe, which is a prior condition in Theorem 1. Finally, the relaxing factor $\kappa_b$ in Lemma~\ref{lemma:relax} must also be selected properly. A smaller choice is suggested if a less conservative control policy is expected, while a larger choice can be made to ensure absolute control safety.

			\textbf{Discussions}.  
			As a prominent feature, our approach can learn an explicit control policy offline and deploy it to a different control scenario even if the concerned constraints are nonlinear and nonconvex. However, in MPC,  the control action must be computed online by periodically solving an optimization problem~\cite{zhang2021robust}, which can be difficult for the on-the-fly implementation under nonlinear and nonconvex constraints, see Section~\ref{sec:app-differ}. As shown in the simulation and real-world experiments,  our learned policy using an inaccurate model shows an impressive sim-to-real transfer capability compared with state-of-the-art model-free RL approaches. In the experiments of differential-drive vehicles, our approach outperforms comparative MPC algorithms under measurement noises and modeling uncertainties.
			Indeed, our approach is a step forward in applying safe RL to the real-world intelligent vehicle control problem. }
		\section{Conclusions}
		This paper proposed a safe RL algorithm with a barrier-based control policy structure and a multi-step policy evaluation mechanism for the optimal control of discrete-time nonlinear systems with time-varying safety constraints.
		{\color{black}Under certain conditions, safety can be guaranteed by our approach in both online and offline learning cases. Our approach can solve continuous control tasks in the environment with abrupt changes, both online and offline. 
			The convergence and robustness of our safe RL under nominal and disturbed scenarios were proven, respectively. The convergence condition of the barrier-based actor-critic learning algorithm was obtained.
			
			Besides numerical simulations for theoretical verification, we tested our approach in two real-world intelligent vehicle platforms. The simulation and real-world experiment results illustrate that our method outperforms state-of-the-art safe RL approaches in control safety, and shows an impressive sim-to-real transfer capability and a satisfactory real-world online learning performance. In general, the proposed safe RL is a step forward in applying safe RL to the optimal control of real-world nonlinear physical systems with time-varying safety constraints. Future works will consider the extension to model-free safe RL with theoretical guarantees.}
		
		\bibliographystyle{unsrt}
		\bibliography{ref}

		\appendix
		\subsection{Proofs of the Main Results}\label{sec:theorem}
		\textbf{\underline{1) Proof of Theorem~\ref{Eqn:theo-con}}} 
		
		(i) 
		First note that,
		\begin{equation*}
		\begin{array}{rl}
		\bar J^{1}(x_k)\hspace{-2mm}&=\sum_{l=0}^{L-1}\gamma^l\bar r(x_{k+l},u^0(x_{k+l}))+ \gamma^L\bar J^0(x_{k+L})\vspace{2mm}\\
		\hspace{-2mm}&=\sum_{l=0}^{L-2}\gamma^l\bar r(x_{k+l},u^0(x_{k+l}))+\vspace{1mm}\\
		\hspace{-2mm}&\quad \quad  \gamma^{L-1}\bar r(x_{k+L-1},u^0(x_{k+L-1}))+\gamma^L\bar J^0(x_{k+L})\vspace{2mm}\\
		\hspace{-2mm}&=\sum_{l=0}^{L-2}\gamma^l\bar r(x_{k+l},u^0(x_{k+l}))+\hspace{-4mm}\quad\min{\{v_{k+L-1},\rho,K\}}\gamma^{L-1}\left\{\right.\vspace{1mm}\\
		\hspace{-2mm}&\left.\bar r(x_{k+L-1},u(x_{k+L-1}))+J^0(x_{k+L})\right\}\vspace{1mm}\\
		\hspace{-2mm}&\leq \sum_{l=0}^{L-2}\gamma^l\bar r(x_{k+l},u^0(x_{k+l}))+\gamma^{L-1}\bar J^0(x_{k+L-1})\vspace{1mm}\\
		\hspace{-2mm}&\vdots\vspace{1mm}\\
		\hspace{-2mm}&\leq \bar r(x_{k},u^0(x_{k}))+\gamma\bar J^0(x_{k+1})=V^0(x_{k})\leq \bar J^0(x_{k}).
		\end{array}
		\end{equation*}
		Hence, $\bar J^{1}(x_k)\leq V^0(x_{k})\leq \bar J^0(x_{k})$. Then one can obtain the result by induction.
		
		
		(iii) Since $\bar J^{i+1}(x_k)\leq \bar J^{i}(x_k)$ and $\bar J^i(x_k)$  is a semi-definite positive function in view of the property of the barrier function. Then one can conclude that $\bar J^{i}(x_k)$ converges to a value denoted as $\bar J^{\infty}(x_k)\geq 0$. From Claim 1), one has $\bar J^{\infty}(x_k)\leq V^{\infty}(x_k)\leq \bar J^{\infty}(x_k)$, then 
		\begin{equation}
		\begin{array}{ll}
		\bar J^{\infty}(x_k)=V^{\infty}(x_k)\\
		=\min{\{v_k,\rho,K\}}\bar r(x_{k},u(x_k))+\gamma\bar J^{\infty}(x_{k+1})\\
		=\min{\{v_{k+i},\rho,K\},i\in\mathbb{N}_0^1}\sum_{l=0}^1 \gamma^l\bar r(x_{k+l},u(x_{k+l}))+\gamma^2\bar J^{\infty}(x_{k+2})\\
		=\hspace{-2mm}\min{\{v_{k+i},\rho,K\},i\in\mathbb{N}_0^{L-1}}\sum_{l=0}^{L-1}\gamma^l\bar r(x_{k+l},u(x_{k+l}))+\gamma^L\bar J^{\infty}(x_{k+L})\\
		=\min{\{v_{k+i},\rho,K\},i\in\mathbb{N}}\bar J^{\infty}(x_k)=\bar J^{\ast}(x_k)
		\end{array}
		\end{equation}
		One can promptly conclude that, $\bar J^{\infty}(x_k)=\bar J^{\ast}(x_k)$. And $v^{\infty}$, $\rho^{\infty}$, and $K^{\infty}$ equal to the optimal values $v^{\ast}(x_k)$, $\rho^{\ast}$, and $K^{\ast}$, respectively. Consequently, $u^{\infty}(x_k)=v^{\infty}(x_k)+\rho^{\infty}\triangledown_v\mathcal{B}_k(v^{\infty}(x_k))+K^{\infty}\triangledown_x\mathcal{B}_k(x_{k})=u^{\ast}(x_k)$.  \hfill $\square$
		
		\textbf{\underline{2) Proof of Proposition~\ref{prop:stable}}} 
		
		In view of~\eqref{Eqn:origin-barrier} and~\eqref{assum:state-con}, one can observe that  $u_k=0$ for $k\geq \bar k$ provided that $v_k=0$ and $x_k=0$. Also, $\mathcal{B}_k(u_k)=0$ as $u_k=0$ and $s_k=0$ as $x_k=0$ for $k\geq \bar k$.  Hence, $\bar J^{\ast}(x_{k})< +\infty$ under $u^{\ast}(x_k)$ with $v^{\ast}(x_k)$, $\rho^{\ast}$, $K^{\ast}$. As a result, $y_k,u_k\rightarrow 0$ as $k\rightarrow +\infty$ under policy $u^{\ast}(x_k)$. Consequently, $x_k\rightarrow 0$ as $k\rightarrow +\infty$. \hfill$\square$
		
		\textbf{\underline{3) Proof of Theorem~\ref{theorem:2}}}
		
		(i) \textbf{Offline learning scenario}. In view of the Lipschitz continuity condition~\eqref{Eqn:lipschitz}, the difference of the real state $z$ under $u^{\ast}(z)$ and the nominal state $x$ under $u^{\ast}(x)$ can be computed as 
		$\|z_{1}-x_{1|0}\|=\|w_0\|\leq \epsilon_w,$ since $x_{0|0}=z_0$.
		Then, by induction, one has 
		$$\|z_{j}-x_{j|0}\|\leq \|z_{j-1}-x_{j-1|0}\|+\epsilon_w\leq \frac{L_f^j-1}{L_f-1}\varepsilon_w.$$
		Hence, the real state $z_k$ converges to $\mathcal{D}_{\varepsilon_w}^{\infty}$ as $k\rightarrow +\infty$.\\
		(ii) \textbf{Online learning scenario}. Given an offline learned control policy, at any time instant $k$, it is learned to obtain an improved control performance under the constraint $x_{k+j|k}\in\bar{\mathcal{X}}_{k+j}$. In this case, the control policy can not be updated if the learned one is evaluated (by MPE) to be inferior to the current one. Given the proof in the offline learning case, the real state of the online learned policy converges to the set $\mathcal{D}_{\varepsilon_w}^{\infty}$.
		\hfill $\square$
		
		\textbf{\underline{4) Proof of Theorem~\ref{THEOM:3}}}
		
		Define a Lyapunov function
		\begin{equation}\label{Eqn:v_{k}}
		\begin{array}{ll}
		V_{k}=\underbrace{{\alpha_c/\gamma_c}\text{tr}(\tilde{W}_{c,k}^{\top}\tilde W_{c,k})}+\underbrace{{1/\gamma_a}\text{tr}(\tilde{W}_{a,k}^{\top}\tilde W_{a,k})}\\
		\hspace{18mm} \alpha_cV_{c,k}\hspace{23mm}V_{a,k}
		\end{array}
		\end{equation}
		where $\alpha_c>1$.
		In view of the update rule~\eqref{Eqn:wc} and~\eqref{Eqn:wa}, one can write the difference of $V_{\star,k}$ ($\star=a,c$ in turns), i.e.,  $\Delta V_{\star,k}=V_{\star,k+1}-V_{\star,k}$ as
		\begin{equation}\label{Eqn:deltaVc(k)}
		\Delta V_{\star,k}=\text{tr}\left(2\tilde{W}_{\star,k}^{\top}\frac{\partial \delta_{\star,k}}{\partial W_{\star,k}}+\gamma_{\star}(\frac{\partial \delta_{\star,k}}{\partial W_{\star,k}})^{\top}\frac{\partial \delta_{\star,k}}{\partial W_{\star,k}}\right)
		\end{equation}
		To first compute $\Delta V_{c}$, note that 
		\begin{equation}\label{Eqn:delta(k)}
		\frac{\partial \delta_{c}}{\partial W_{c}}=\frac{\partial <\varepsilon_{c},\varepsilon_{c}>}{\partial W_c}
		=2\Delta h_c\varepsilon_{c}^{\top}.
		\end{equation}
		Moreover, 
		\begin{equation}\label{Eqn:epsilonc(k)}
		\begin{array}{ll}
		\varepsilon_{c,k}=\sum_{l=0}^{L-1}\gamma^l\bar r(x_{k+l},u_{k+l})+ \gamma^L\hat {\bar J}(x_{k+L})-\hat {\bar J}(x_{k})\\
		= -\gamma^L{\bar J}^{\ast}(x_{k+L})+ {\bar J}^{\ast}(x_{k})+\gamma^L\hat {\bar J}(x_{k+L})-\hat {\bar J}(x_{k})\\
		=-\tilde{W}_{c,k}^{\top}\Delta h_{c,k}-\Delta \kappa_{c,k}
		\end{array}
		\end{equation}
		where $\Delta \kappa_{c,k}$=$\gamma^L\kappa_{c,k+L}-\kappa_{c,k}$, the second equality in~\eqref{Eqn:epsilonc(k)} is due to the Bellman equation~\eqref{Eqn:V_bzk}. 
		
		Taking~\eqref{Eqn:delta(k)} with~\eqref{Eqn:epsilonc(k)} into~\eqref{Eqn:deltaVc(k)}, leads to
		\begin{equation}\label{Eqn:delta_Vc_final(k)}
		\begin{array}{lll}
		\hspace{-2mm}\Delta V_{c}
		&\hspace{-2mm}=&\hspace{-2mm}-4c_1\|\tilde W_c\|^2+4\Delta \kappa_cc_2\tilde W_c+4\gamma_{c}\Delta\bar h_c\Delta \kappa_c^2\\
		\hspace{-2mm}&\hspace{-2mm}\leq&\hspace{-2mm}-4(c_1-\lambda_{\rm max}(\bar c_2)\beta_1)\|\tilde W_c\|^2+\epsilon_{\kappa,1}
		
		\end{array}
		\end{equation}
		where $c_1=(1-\gamma_{c}\Delta\bar h_c)\Delta\bar h_c$, $c_2=-(1-2\gamma_{c}\Delta\bar h_c)\Delta h_c^{\top}$, $\bar c_2=c_2^{\top}c_2$, $\epsilon_{\kappa,1}=4(\gamma_{c}\Delta\bar h_c+1/\beta_1)\Delta \kappa_c^2$, $\beta_1$ is a tuning constant.
		
		To  compute $\Delta V_{a}$, inline with~\eqref{Eqn:delta(k)}, one has
		\begin{subequations}\label{Eqn:delta_wA}
			\begin{align}
			\frac{\partial \delta_{a}}{\partial \hat{\rho}}=-2\triangledown\mathcal{B}(v)\varepsilon_{a}^{\top}\bar R
			\end{align}
			where $\bar R=2R+\mu\triangledown^2\mathcal{B}(u)$, $\triangledown\mathcal{B}(z)$ stands for $\triangledown_z\mathcal{B}(z)$ for a general variable $z$;
			and 
			\begin{align}
			\begin{array}{ll}
			\frac{\partial \delta_{a}}{\partial W_{a1}}
			=-2h_{a1}\varepsilon_{a}^{\top}\bar R.
			\end{array}
			\end{align}
		\end{subequations}
		
		With~\eqref{Eqn:delta_wA}, we write $\Delta V_a=  \Delta V_{a1}+  \Delta V_{a2}$ where $\Delta V_{a1}$ and $\Delta V_{a2}$ are given as
		\begin{subequations}\label{Eqn:delta_Va_final(k)}
			\begin{align}
			\hspace{-1.5mm}	\Delta V_{a1}&=4\text{tr}(-\tilde{W}_{a1}^{\top}h_{a1}\varepsilon_{a}^{\top}\bar R+\gamma_{a}\bar R^{\top}\varepsilon_{a}\bar h_{a1}\varepsilon_{a}^{\top}\bar R)\label{Eqn:delta_Va1_final(k)}\\
			\hspace{-1.5mm}	\Delta V_{a2}&=4\text{tr}(-\tilde {\rho}\triangledown\mathcal{B}(v)\varepsilon_{a}^{\top}\bar R+\gamma_{a}\bar R^{\top}\varepsilon_{a}\bar {\mathcal{B}}_v\varepsilon_{a}^{\top}\bar R),\label{Eqn:delta_Va2_final(k)}
			\end{align}
		\end{subequations}
		where $\tilde {\rho}=\rho-\hat \rho$, $\bar h_{a1}=h_{a1}^{\top}h_{a1}$, $\bar {\mathcal{B}}_v=\triangledown\mathcal{B}(v)^{\top}\triangledown\mathcal{B}(v)$.
		In view of~\eqref{Eqn:delta_Va_final(k)}, letting $\bar h_a=\bar h_{a1}+\bar {\mathcal{B}}_v$, one can write $\Delta V_a$ as
		\begin{equation}\label{Eqn:delta-Va}
		\begin{array}{lll}
		\Delta V_a&=&  \Delta V_{a1}+  \Delta V_{a2}\\
		&=&-4\text{tr}(\xi_a\varepsilon_{a}^{\top}\bar R)+4\gamma_{a}\text{tr}(\bar R^{\top}\varepsilon_{a}\bar h_a\varepsilon_{a}^{\top}\bar R).
		\end{array}
		\end{equation}
		Let $g_c=\triangledown_u f(x,u)^{\top}(\triangledown h_c^+)^{\top}$,  $\bar \kappa_1=\triangledown_u f(x,u)^{\top}\triangledown\kappa_c^++2R\kappa_a$, then
		\begin{equation*}
		\begin{array}{lll}
		\varepsilon_{a}&=&\nu^d-\nu^{\ast}+\nu^{\ast}-\nu\\
		&=&\nu^d+g_cW_c^{\ast}+\bar \kappa_1+2R(W_a^{\ast})^{\top}h_a+\\
		&&\mu\triangledown\mathcal{B}(u^{\ast})-2RW_a^{\top}h_a-\mu\triangledown\mathcal{B}(u)
		\end{array}
		\end{equation*}
		In view of the definition of $\mathcal{B}$ and $G^{i}(u)=E^iu$, it holds that
		\begin{equation}\label{Eqn:barrier-diff}
		\bar{\mathcal{B}}=\triangledown\mathcal{B}(u^{\ast})-\triangledown \mathcal{B}(u)=-D(\xi_a+\kappa_a)
		\end{equation}
		where
		$D=\sum_{i=1}^{p_u}(E^i)^{\top}E^i/(G^i(u^{\ast})G^i(u))$ for $\bar\sigma\geq\kappa_b$ and $D=2H_u$ otherwise.
		Hence, with~\eqref{Eqn:barrier-diff}, one has
		\begin{equation}\label{Eqn:epsilona(k)}
		\begin{array}{lll}
		\varepsilon_{a}=g_c\tilde W_c+R_d\xi_a+\bar \kappa.
		\end{array}
		\end{equation}
		where $R_d=2R-\mu D$ and $\bar \kappa=\bar \kappa_1-\mu D\kappa_a$.
		Taking~\eqref{Eqn:epsilona(k)} into~\eqref{Eqn:delta-Va},  one can compute:
		\begin{equation}\label{Eqn:delta-Va-final0}
		\begin{array}{lll}
		\Delta V_a
		=4(-\|\xi_a\|_{R_d\bar R}^2-\xi_a^{\top}\bar Rg_c\tilde W_c-\xi_a^{\top}\bar R\bar\kappa)+d\|\varepsilon_{a}\|_{\bar R^2}^2
		\end{array}
		\end{equation}
		where $d=4\gamma_{a}\bar h_a$.
		Applying Young's inequality to~\eqref{Eqn:epsilona(k)}, leads to
		$\|\varepsilon_{a}\|^2\leq 3\|\tilde W_c\|_{\bar g_c}^2+3\|\xi_a\|_{R_d^2}^2+3\|\bar \kappa\|^2$,
		where $\bar g_c=g_c^{\top} g_c$.
		Hence, one can conclude from~\eqref{Eqn:delta-Va-final0} that 
		\begin{equation}\label{Eqn:DELTAVa-final}
		\begin{array}{lll}
		\Delta V_a\leq -\|\xi_a\|_P^2+(4/\beta_2+3d)\lambda_{\rm max}({\bar g_c})\|\tilde W_c\|^2+\epsilon_{\kappa,2},
		\end{array}
		\end{equation}
		where $P=4R_d\bar R-(\beta_2+\beta_3)\bar R^2-3d R_d^2\bar R^2,$
		$\beta_2$, $\beta_3$ are tuning constants,  $\epsilon_{\kappa,2}=(4/\beta_3+3d)\|\bar \kappa\|_{\bar R^2}^2$.
		Therefore, combining~\eqref{Eqn:delta_Vc_final(k)} and~\eqref{Eqn:DELTAVa-final}, it holds that
		\begin{equation}\label{Eqn:DELTAV-final}
		\begin{array}{lll}
		\Delta V
		&\leq& -\|(\xi_a,\tilde W_c)\|_S^2+\epsilon_{t}
		\end{array}
		\end{equation}
		where  $\epsilon_{t}=\alpha_c\epsilon_{\kappa,1}+\epsilon_{\kappa,2}$, matrix $S$ is given as
		\begin{equation}\label{Eqn:definite-matrix}
		\begin{array}{ll}
		S=\text{diag}\{P, 4\alpha_c(c_1-\bar c_2\beta_1)-(4/\beta_2+3d)\lambda_{\rm max}({\bar g_c})
		\}.
		\end{array}
		\end{equation}
		Note that, in view of Assumption~\ref{assum:f} and Lemma~\ref{lemma:relax}, $\lambda_{\rm max}({\bar g_c})$ is bounded by 
		$\lambda_{\rm max}({\bar g_c})\leq \|\triangledown_u f(x,u)^{\top}(\triangledown h^{+}_c)\|^2\leq g_m^2(\bar\sigma_{c,m}^2+\mathcal{B}_{x,m}^2),$
		$ R_d,\,\bar R\preccurlyeq 2R+2\mu H_u,
		$
		and   $d$ can be made small by tuing $\gamma_a$, i.e.,
		$d\leq 4\gamma_{a}\|h_a\|^2\leq 4\gamma_{a}(\sigma_{a,m}^2+\mathcal{B}_{v,m}^2+\mathcal{B}_{x,m}^2):=d_m.
		$
		Hence,  thanks to~\eqref{Eqn:actor-cond}, one can first tune $\beta_2$ and $\beta_3$ such that $P\succ 0$. Then, in view of~\eqref{Eqn:critic-cond}, one can  tune $\alpha_c$ and  $\beta_1$ to make sure the second term of~\eqref{Eqn:definite-matrix} is positive-definite.
		
		Since $\epsilon_t$ is bounded, let $\epsilon_{m}$ be the lower bound of $\epsilon_t$; then in view of Assumption~\ref{assum:network}, with $S\succ 0$, it follows that $\Delta V\leq 0$ for all 
		$ \|(\xi_{a,k},\tilde W_{c,k})\|\geq \sqrt{\frac{\epsilon_{m}}{\lambda_{\rm min}(S)}}.$ Consequently,
		$ \|(\xi_{a,k},\tilde W_{c,k})\|\rightarrow 0,\ \text{as}\, k\rightarrow +\infty,$ provided that $\epsilon_{t}\rightarrow 0$.  \hfill $\square$
					\begin{table*}[h!tb]
			\centering \caption{{\color{black}Numerical comparisons under dynamic obstacles. Cost $J_e=1/M\sum_{j=1}^M\|e_j\|^2$. $d_r$ represents the distance between adjacent reference points. ``S" and ``F" stand for ``Succeed" and ``Fail", respectively.}}
			\label{tab:Tab_com-suc0}
			\renewcommand\arraystretch{1.1}
			\scalebox{0.8}{
				\begin{tabular}{ccccccc}
					\toprule
					\multirow{2}{*}{Methods}&\multicolumn{3}{c}{Scenarios}&\multirow{2}{*}{Coll. avoid./overtak.}&\multirow{2}{*}{$J_e$ (in coll. avoid.)}&\multirow{2}{*}{$J_e$ (in path foll.)}\\\cline{2-4}
					&Parameters&$d_o$(m)&$d_r$(m)&\multicolumn{3}{c}{}\\\hline
					\multirow{3}{*}{Ours}&\multirow{3}{*}{--}&\multirow{3}{*}{--}&$0.07$& S/S&\textbf{0.237}&\textbf{0.003}\\\cline{4-7}
					&&&$0.56$&S/S&{\textbf{0.286}}&0.031\\\cline{4-7}
					&&&$1.12$&S/S&\textbf{0.365}&\textbf{0.107}\\\midrule
					\multirow{6}{*}{NMPC-c}&\multirow{3}{*}{$\mu_p=5\cdot 10^{-4}$}&\multirow{3}{*}{1}&$0.07$&S/F&--&--\\\cline{4-7}
					&&&$0.56$ &S/S&{0.579}&0.03\\\cline{4-7}
					&&&$1.12$ &S/S&0.96&0.199\\\cline{2-7}
					&\multirow{3}{*}{$\mu_p=5\cdot 10^{-3}$}&\multirow{3}{*}{1}&$0.07$&S/F&--&--\\\cline{4-7}
					&&&$0.56$ &S/S&{0.452}&0.03\\\cline{4-7}
					&&&$1.12$ &S/S&0.639&0.139\\\midrule
					\multirow{6}{*}{NMPC-e}&\multirow{3}{*}{$\mu_p=5\cdot 10^{-4}$}&\multirow{3}{*}{1}&$0.07$&S/F&--&--\\\cline{4-7}
					&&&$0.56$&S/S&0.62&\textbf{0.029}\\\cline{4-7}
					&&&$1.12$&S/S&0.782&0.158\\\cline{2-7}
					&\multirow{3}{*}{$\mu_p=5\cdot 10^{-3}$}&\multirow{3}{*}{1}&$0.07$&S/F&--&--\\\cline{4-7}
					&&&$0.56$&S/S&0.885&0.033\\\cline{4-7}
					&&&$1.12$&S/S&0.798&0.164\\
					\midrule
					\multirow{36}{*}{NMPC-cbf}
					&\multirow{3}{*}{$\eta=0.4$}&\multirow{3}{*}{1}&$0.07$&S/F&--&--\\\cline{4-7}
					&&&$0.56$&S/S&0.669&0.031\\\cline{4-7}
					&&&$1.12$&S/S&0.798&0.164\\\cline{2-7}
					&\multirow{3}{*}{$\eta=0.5$}&\multirow{3}{*}{1}&$0.07$&S/F&--&--\\\cline{4-7}
					&&&$0.56$&S/S&0.618&0.03\\\cline{4-7}
					&&&$1.12$&S/S&1.12&0.118\\\cline{2-7}
					&\multirow{3}{*}{$\eta=0.6$}&\multirow{3}{*}{1}&$0.07$&S/F&--&--\\\cline{4-7}
					&&&$0.56$&S/S&0.717&0.05\\\cline{4-7}
					&&&$1.12$&S/S&1.254&0.273\\\cline{2-7}
					&\multirow{3}{*}{$\eta=0.8$}&\multirow{3}{*}{1}&$0.07$&S/F&--&--\\\cline{4-7}
					&&&$0.56$&S/S&1.048&0.03\\\cline{4-7}
					&&&$1.12$&S/S&0.944&0.35\\\cline{2-7}
					&\multirow{3}{*}{$\eta=1.0$}&\multirow{3}{*}{1}&$0.07$&S/F&--&--\\\cline{4-7}
					&&&$0.56$&S/S&0.589&0.03\\\cline{4-7}
					&&&$1.12$&S/S&0.961&0.139\\\cline{2-7}
					&\multirow{2}{*}{$\eta=2.5$}&\multirow{2}{*}{1}
					&$0.56$&S/S&0.428&0.06\\\cline{4-7}
					&&&$1.12$&S/S&0.626&0.3\\\cline{2-7}
					&\multirow{4}{*}{$\eta=5.0$}&\multirow{2}{*}{1}
					&$0.56$&S/S&0.772&0.05\\\cline{4-7}
					&&&$1.12$&S/S&0.951&0.45\\\cline{3-7}
					&&\multirow{2}{*}{1.1}&$0.56$&S/S&1.274&0.04\\\cline{4-7}
					&&&$1.12$&S/S&1.277&0.45\\\cline{2-7}
					&\multirow{4}{*}{$\eta=7.5$}&\multirow{2}{*}{1}
					&$0.56$&S/S&0.808&0.055\\\cline{4-7}
					&&&$1.12$&S/S&0.644&0.116\\\cline{3-7}
					&&\multirow{2}{*}{1.1}&$0.56$&S/S&0.947&0.04\\\cline{4-7}
					&&&$1.12$&S/S&1.377&0.84\\\cline{2-7}
					&\multirow{4}{*}{$\eta=10$}&\multirow{2}{*}{1}
					&$0.56$&S/S&1.136&0.1\\\cline{4-7}
					&&&$1.12$&S/S&0.781&0.36\\\cline{3-7}
					&&\multirow{2}{*}{1.1}&$0.56$&S/S&1.359&0.07\\\cline{4-7}
					&&&$1.12$&S/S&1.352&0.789\\\cline{2-7}
					&\multirow{4}{*}{$\eta=12.5$}&\multirow{2}{*}{1}
					&$0.56$&S/S&1.623&0.05\\\cline{4-7}
					&&&$1.12$&S/S&1.384&0.567\\\cline{3-7}
					&&\multirow{2}{*}{1.1}&$0.56$&S/S&0.428&0.06\\\cline{4-7}
					&&&$1.12$&S/S&0.626&0.3\\\cline{2-7}
					&\multirow{3}{*}{$\eta=15$}&\multirow{2}{*}{1}
					&$0.56$&S/S&0.58&0.04\\\cline{4-7}
					&&&$1.12$&S/S&1.11&0.549\\\cline{3-7}
					&&\multirow{1}{*}{1.1}&$0.56$&S/S&0.577&0.04\\
					\bottomrule
				\end{tabular}
			}
		\end{table*}
		\begin{figure}[H]
			\centering
			\includegraphics[width=0.5\textwidth]{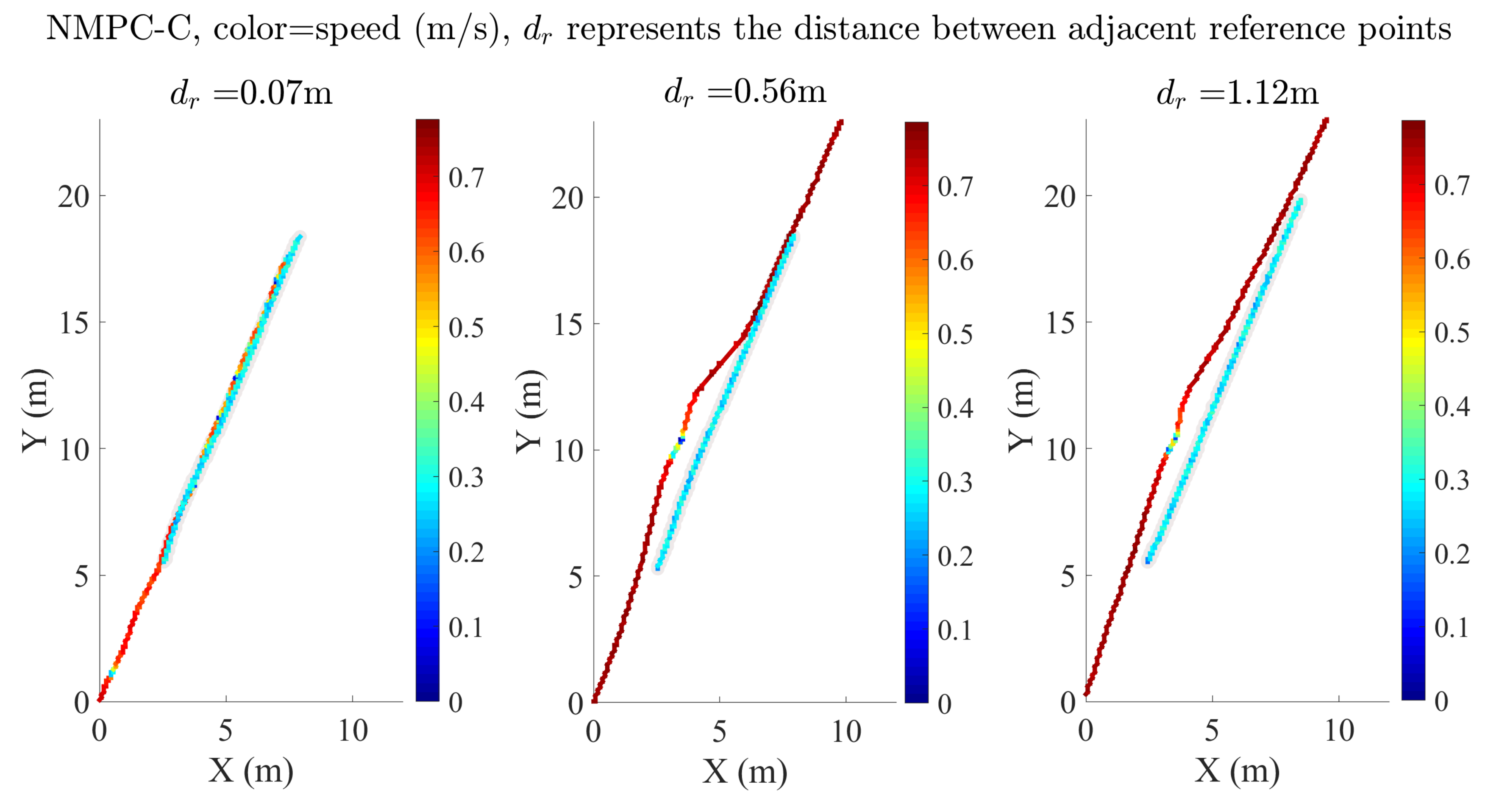}
			\caption{{\color{black}The experimental results on path-following and collision avoidance by NMPC-c ($\mu_p=5\cdot 10^{-3}$): the shorter line with a gray shade represents the route of the moving vehicle, while the longer line represents the trajectory of the ego vehicle. When using dense reference points ($d_r=0.07$m), the ego vehicle was unable to pass, but was successful when using sparse reference points. In the latter scenario, the collision avoidance process caused the ego vehicle to experience a short transient period of rapid speed variation.}} 
			\label{fig:nmpcc}
		\end{figure}
		\begin{figure}[H]
			\centering
			\includegraphics[width=0.5\textwidth]{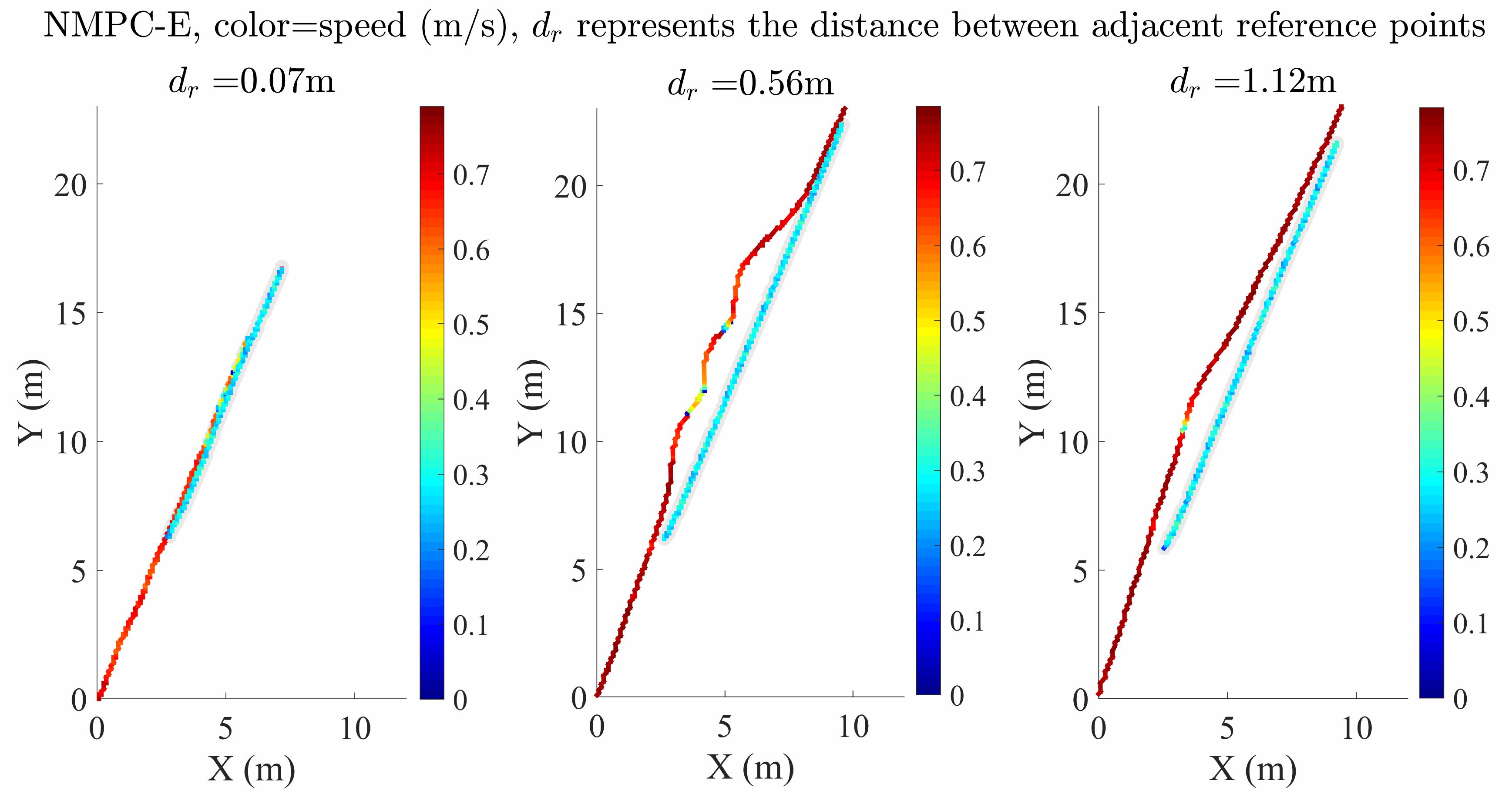}
			\caption{{\color{black}The experimental results on path-following and collision avoidance by NMPC-e ($\mu_p=5\cdot 10^{-4}$): the shorter line with a gray shade represents the route of the moving vehicle, while the longer line represents the trajectory of the ego vehicle. When using dense reference points ($d_r=0.07$m), the ego vehicle was unable to pass, but was successful when using sparse reference points. In the latter scenario, the collision avoidance process caused the ego vehicle to experience a short transient period of rapid speed variation.}} 
			\label{fig:nmpce}
		\end{figure}
		\begin{figure}[H]
			\centering
			\includegraphics[width=0.5\textwidth]{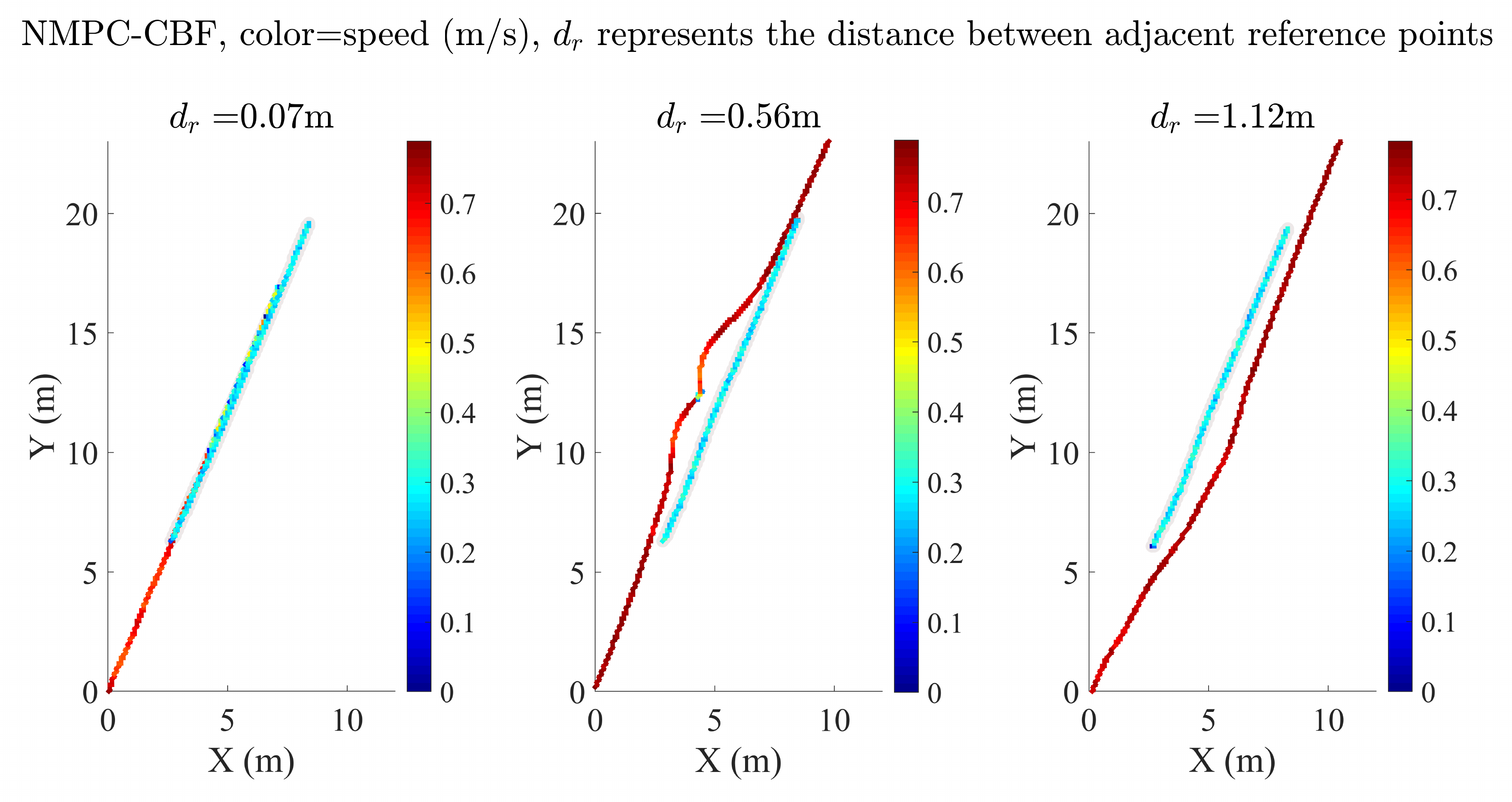}
			\caption{{\color{black}The experimental results on path-following and collision avoidance by NMPC-cbf ($\eta=0.5$): the shorter line with a gray shade represents the route of the moving vehicle, while the longer line represents the trajectory of the ego vehicle. When using dense reference points ($d_r=0.07$m), the ego vehicle was unable to pass, but was successful when using sparse reference points. In the latter scenario, the collision avoidance process caused the ego vehicle to experience a short transient period of rapid speed variation.}} 
			\label{fig:nmpccbf}
		\end{figure}
			\begin{figure}[H]
		\centering
		\includegraphics[width=0.5\textwidth]{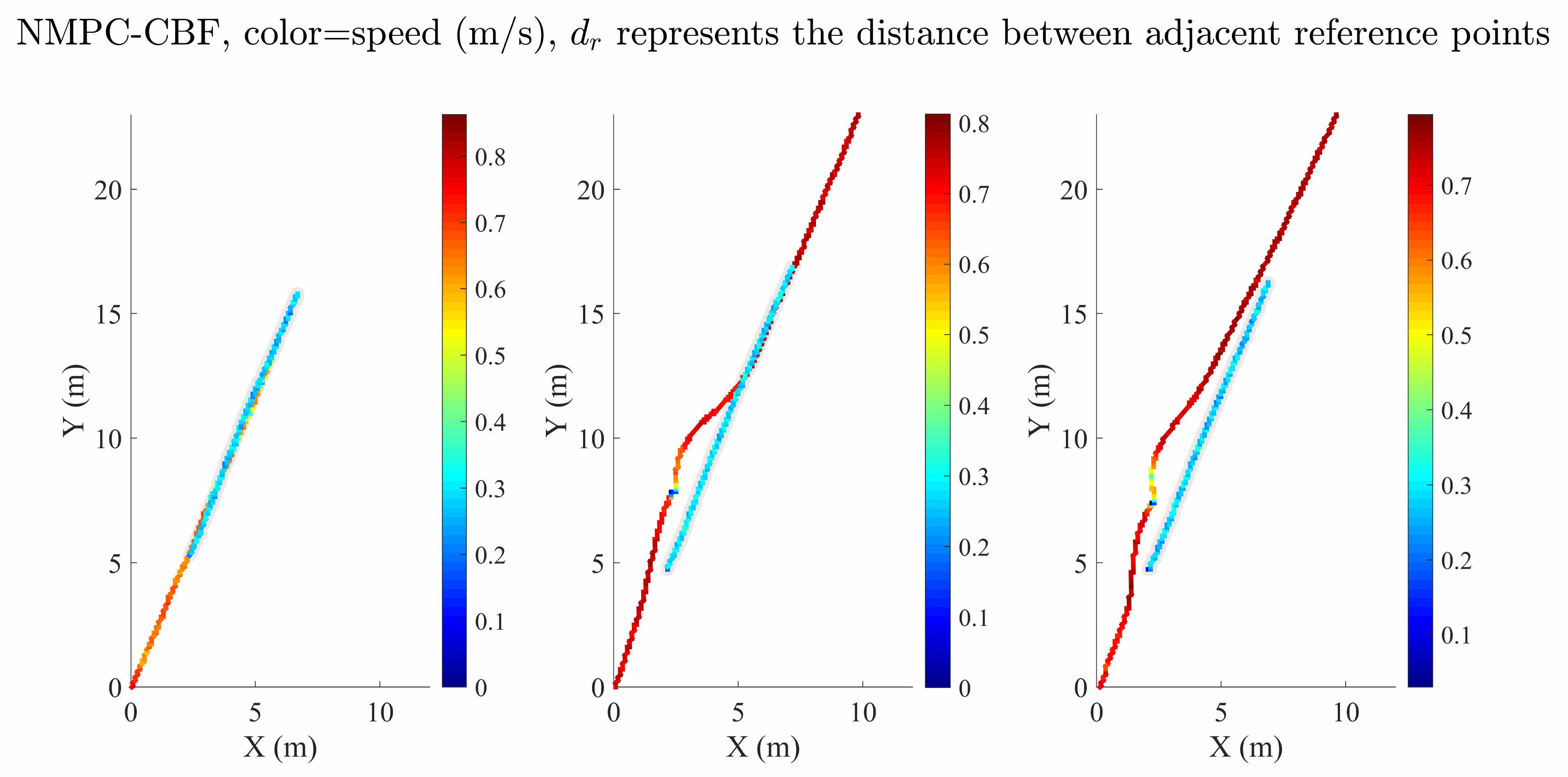}
		\caption{{\color{black}The experimental results on path-following and collision avoidance by NMPC-cbf ($\eta=1$): the shorter line with a gray shade represents the route of the moving vehicle, while the longer line represents the trajectory of the ego vehicle. When using dense reference points ($d_r=0.07$m), the ego vehicle was unable to pass, but was successful when using sparse reference points. In the latter scenario, the collision avoidance process caused the ego vehicle to experience a short transient period of rapid speed variation.}} 
		\label{fig:nmpccbf1}
	\end{figure}
		\begin{figure}[H]
			\centering
			\includegraphics[width=0.5\textwidth]{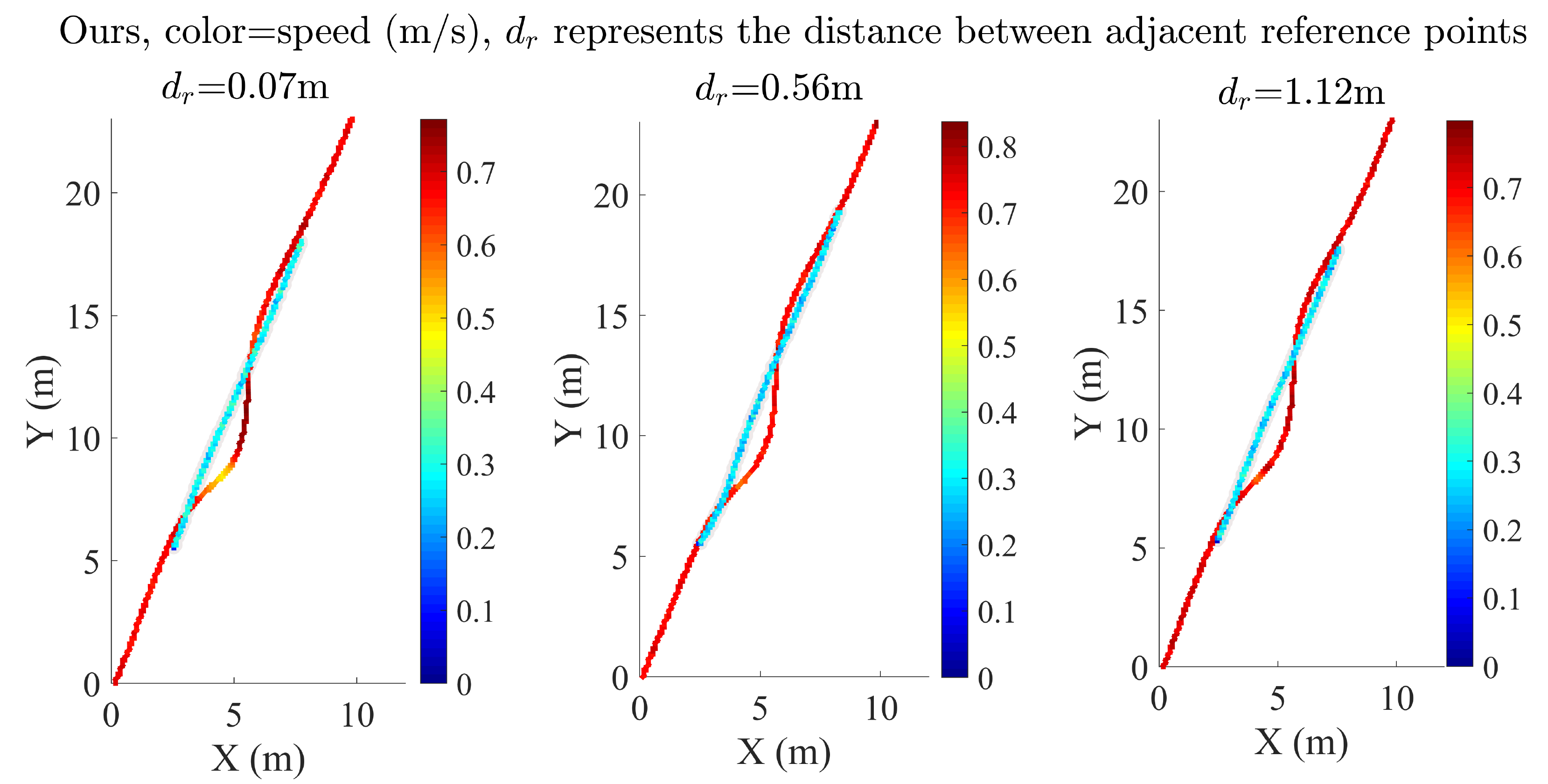}
			\caption{{\color{black}The experimental results on path-following and collision avoidance by our approach: the shorter line with a gray shade represents the route of the moving vehicle, while the longer line represents the trajectory of the ego vehicle. The ego vehicle avoided and overtook the moving vehicle in all scenarios.}} 
			\label{fig:our}
		\end{figure}
		\begin{figure}[H]
			\centering
			\includegraphics[width=0.5\textwidth]{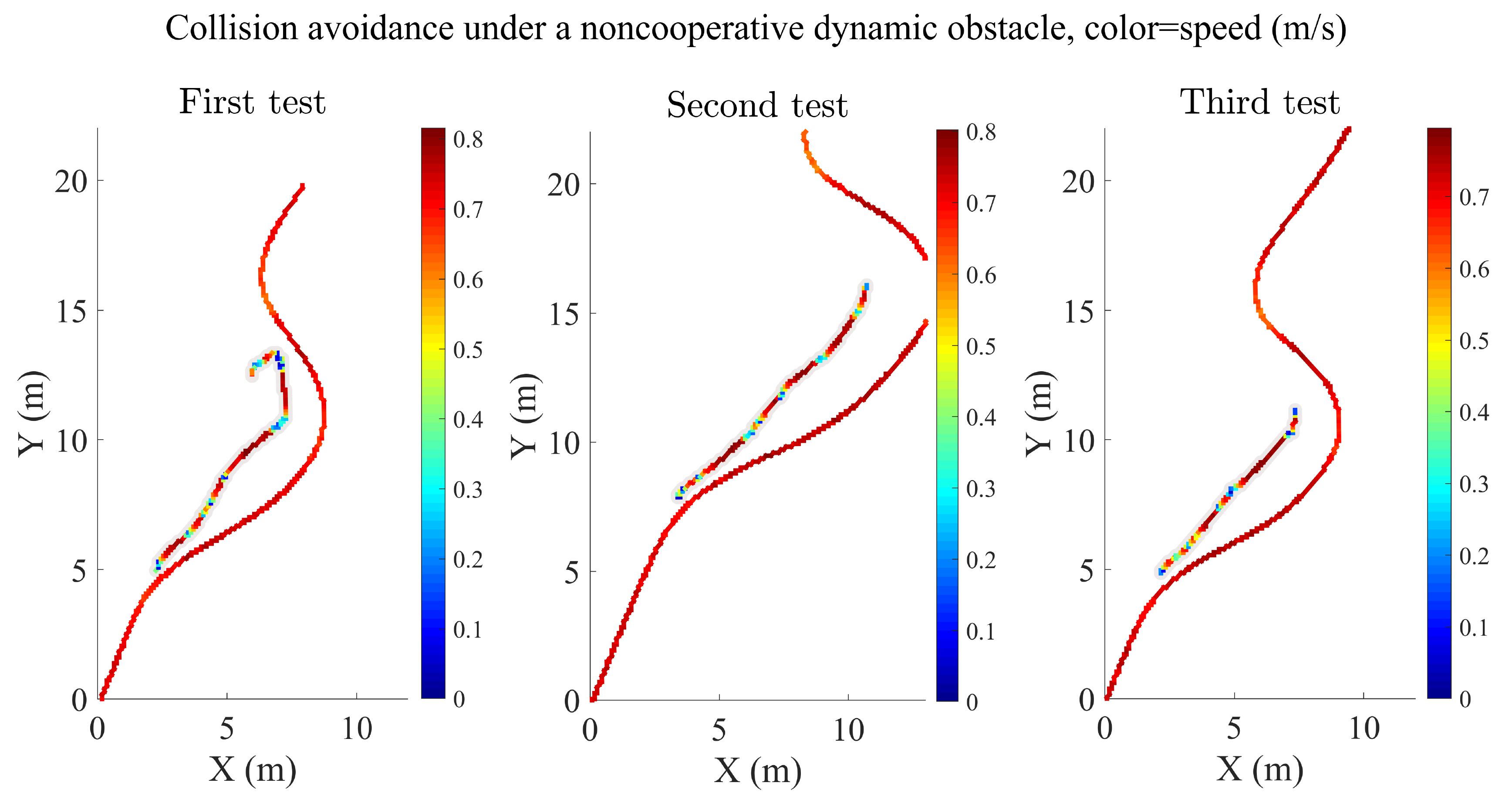}
			\caption{{\color{black}Noncooperative dynamic collision avoidance by our approach. The moving vehicle (obstacle) was manipulated by a human remote control handle to block the path, when the ego vehicle actively avoided collision with the obstacle.}} 
			\label{fig:our-noncoor}
		\end{figure}
		\begin{figure}[H]
			\centering
			\includegraphics[width=0.4\textwidth]{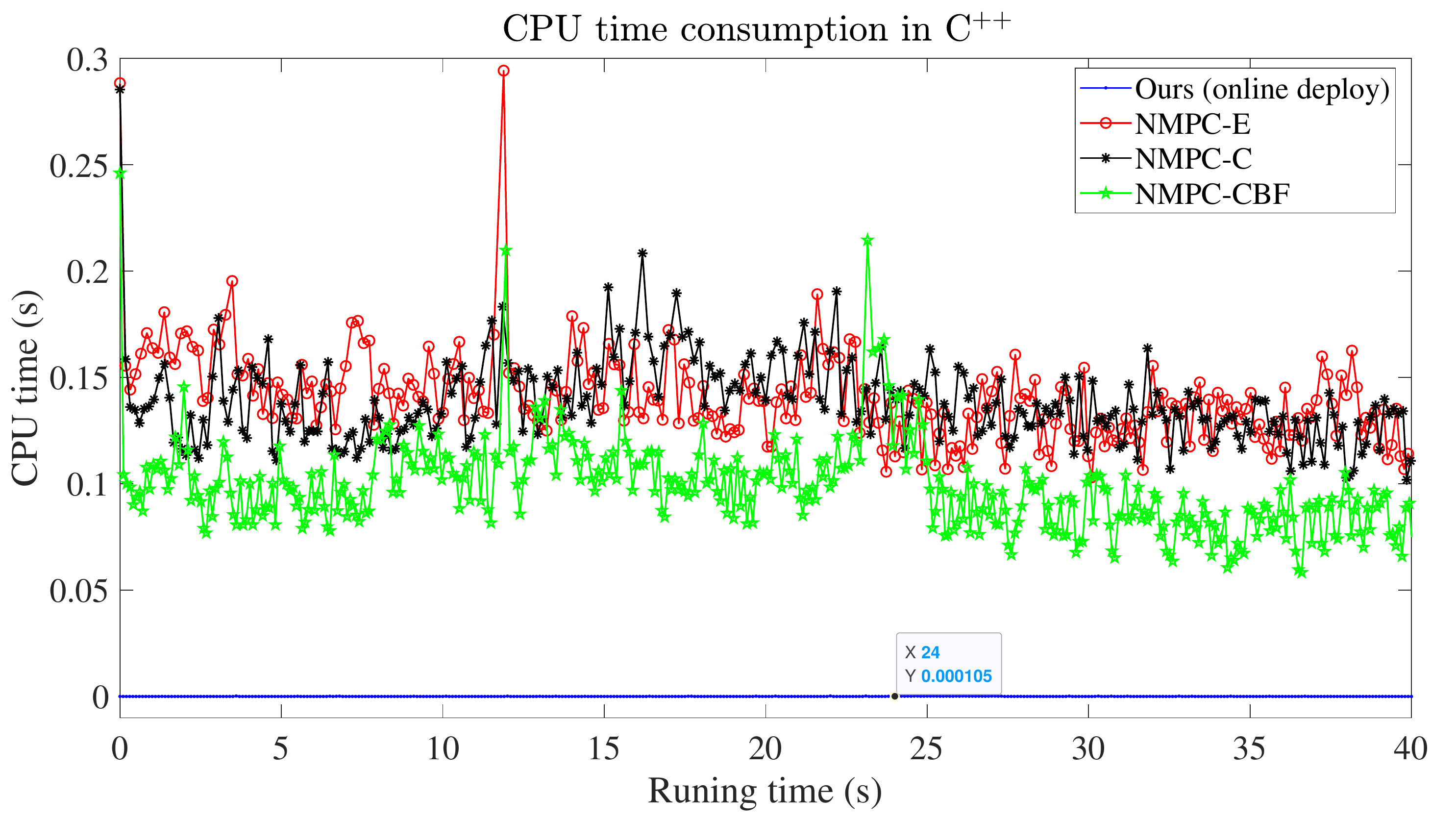}
			\caption{{\color{black}The CPU running time comparison in C$^{++}$. In many time instants, the computational time values of the NMPC-c, NMPC-e, NMPC-cbf are greater than the adopted sampling interval, i.e., 0.1s, which could hamper the control performance (see Table~\ref{tab:Tab_com-suc0}), while the computational time value of our approach is much smaller and its influence on the control performance can be negligible.}} %
			\label{fig:cputime}
		\end{figure}

		{\color{black}	\subsection{Auxiliary Real-World Experimental Results on the Differential-Drive Vehicle}\label{appen:experi-results} 
			Figs.~\ref{fig:nmpcc}-\ref{fig:our} present the experimental results of NMPC-c, NMPC-e, NMPC-e, and our approach to path following with dynamic collision avoidance. Fig.~\ref{fig:our-noncoor} presents the experimental results of our approach in a noncooperative dynamic collision avoidance scenario, and Fig.~\ref{fig:cputime} shows the significant advantage of our approach in computational load reduction in the C$^{++}$ environment.}
		\subsection{Simulation on Regulation of a Mass-Point Robot}\label{sec:simulation} Consider the regulation control of a mass-point robot. Its discrete-time model is described by $x_{k+1}=Ax_k+Bu_k$
		where 
		$$A=\begin{bmatrix}
		0.995& 0.0998\\-0.0998& 0.995
		\end{bmatrix}, B=\begin{bmatrix}
		-0.2\\-0.1
		\end{bmatrix},$$
		$x=(x_1,x_2)$ are the positions to the references. 
		\begin{figure*}[h]
			\centering
			\includegraphics[width=0.75\textwidth]{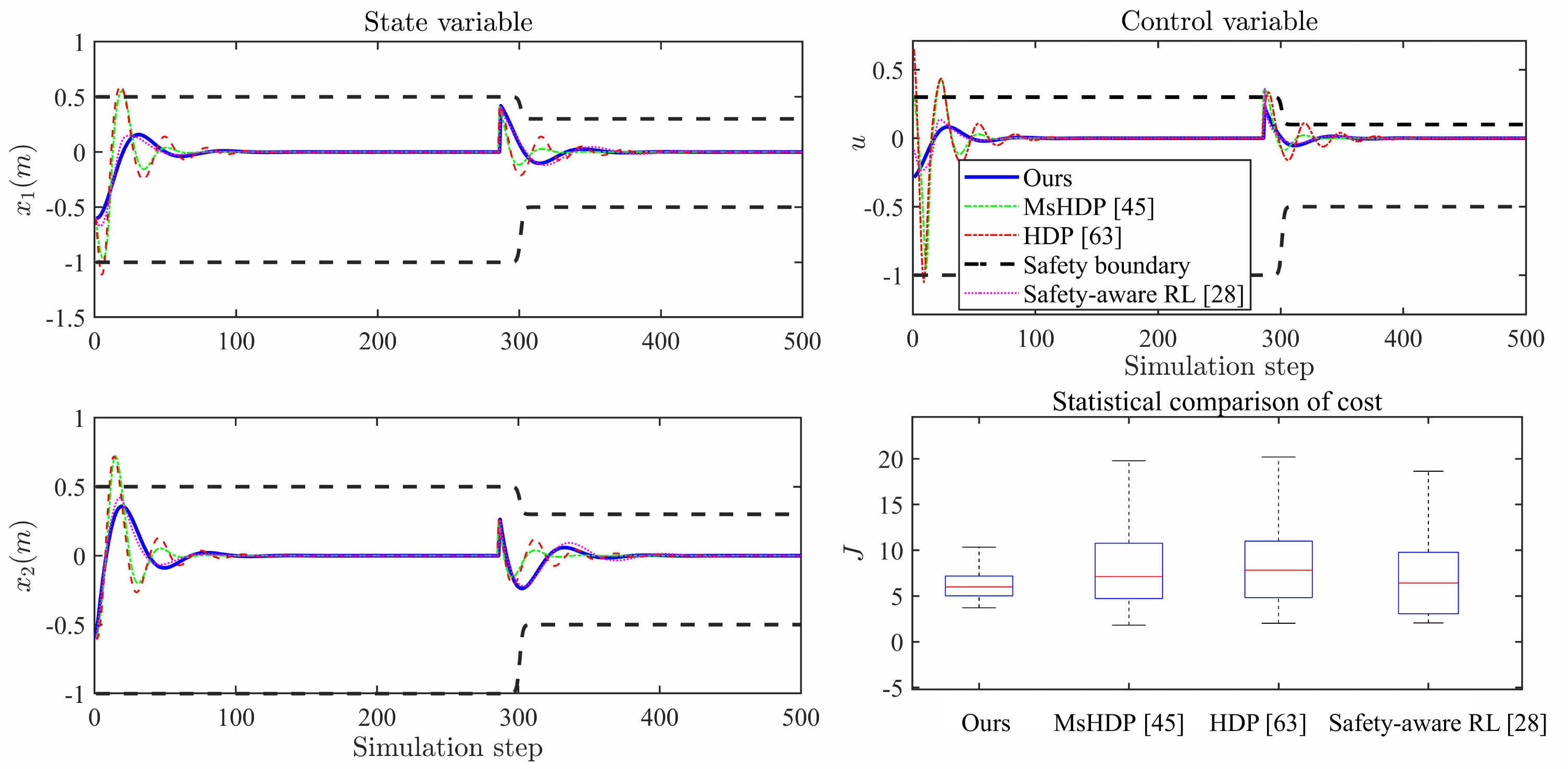}
			\caption{Simulation results for control of mass-point robot: The state, control, and cost comparisons between our approach and algorithms in~\cite{yang2019safety},~\cite{luo2017output}, and~\cite{zhang2008novel}. Right-bottom panel: Statistical comparison of cost with 500 repetitive experiments; red lines stand for mean value of $J$, black boxes are the area of standard deviations of $J$, while  black horizontal lines are the maximal and minimal deviations.}
			\label{fig:com-state}
		\end{figure*}
		\begin{table}[ht]
			\centering \caption{Comparisons in terms of safety rate in mass-point robot example. }
			\label{tab:Tab_com0}
			\vskip 0.1cm
			\scalebox{0.75}{
				\begin{tabular}{@{}ccc@{}}
					\toprule
					Approach          & Ours &  Safety-aware RL~\cite{yang2019safety}\\\midrule
					Time-invariant constraint             & 100\% & 100\%                   \\ 
					Time-varying constraint             & 100\% & $\leq$ 95\% \\   
					\bottomrule
			\end{tabular}}
		\end{table}
		
		In the simulation process, the  penalty matrices were selected as $Q=I$, $R=0.1$, $\mu=0.001$. The discounting factor $\gamma$ was $\gamma=0.95$. The step $L$ was chosen as $L=10$. The entries of weighting matrices $W_c$ and $W_a$  were initialized with uniformly random numbers. The initial state was $x=(-0.5,-0.5)$. The state and control were initially limited by 
		$-1\leq x_1,x_2\leq 0.5,\quad -1\leq u\leq 0.3. 
		$ Then at $k=285$, $x$ was reset as $x=(-0.65,-0.65)$ and the constraints were changed as $-0.5\leq x_1,x_2\leq 0.3,\, -0.5\leq u\leq 0.1$. 
		
		The control performances are compared in Fig.~\ref{fig:com-state}, which shows that all approaches can converge to the origin. In contrast, our approach and safety-aware RL~\cite{yang2019safety} can ensure safety constraint satisfaction in the control process. 
		Moreover, comparisons in terms of safety for 500 repetitive experimental tests are listed in Table~\ref{tab:Tab_com0}, which illustrates that our approach can ensure safety for all the performed tests, while it did not hold so for safety-aware RL in~\cite{yang2019safety} under time-varying safety constraints. The reason behind this is the adopted multi-step policy evaluation mechanism in our approach. From the box-plot in the right-bottom panel of Fig.~\ref{fig:com-state}, one can see that the mean values of $J$ in all approaches are comparable and its 
		standard cost deviation in our approach is smaller than that in~\cite{yang2019safety},~\cite{luo2017output}, and~\cite{zhang2008novel}. The results reveal that the performance of our approach is more stable under state and control constraints due to the proposed barrier-based control policy design.
		\subsection{Simulation on Regulation of Van Der Pol Oscillator} 
		Consider the regulation control of Van der Pol oscillator~\cite{zhang2021robust}. Its discrete-time model is given as
		\begin{equation*}\label{Eqn:ct}
		\left\{\begin{array}{ll}
		x_{1,k+1}=x_{1,k}+\Delta t x_{2,k}\\
		x_{2,k+1}
		=x_{2,k}+\Delta t(x_{2,k}-x_{1,k}^2x_{2,k}-x_{1,k}+u_k)
		\end{array}\right.
		\end{equation*}
		where $x_1$ and $x_2$ are the states, and $u$ is the control variable, $\Delta t=0.01$s. In the simulation process, the  penalty matrices were selected as $Q=I$, $R=0.1$, $\mu=0.01$. The discounting factor $\gamma$ was $\gamma=0.95$. The step $L$ was chosen as $L=10$. The entries of weighting matrices $W_c$ and $W_a$  were initialized using saturated uniformly random numbers such that Theorem~\ref{Eqn:theo-con} was fulfilled. Starting with an initial condition $x_0=(-0.5,-0.5)$, the training was performed under time-varying state constraints, see Fig.~\ref{fig:com-cost-van}. 
		\begin{figure*}[h]
			\centering
			\includegraphics[width=0.6\textwidth]{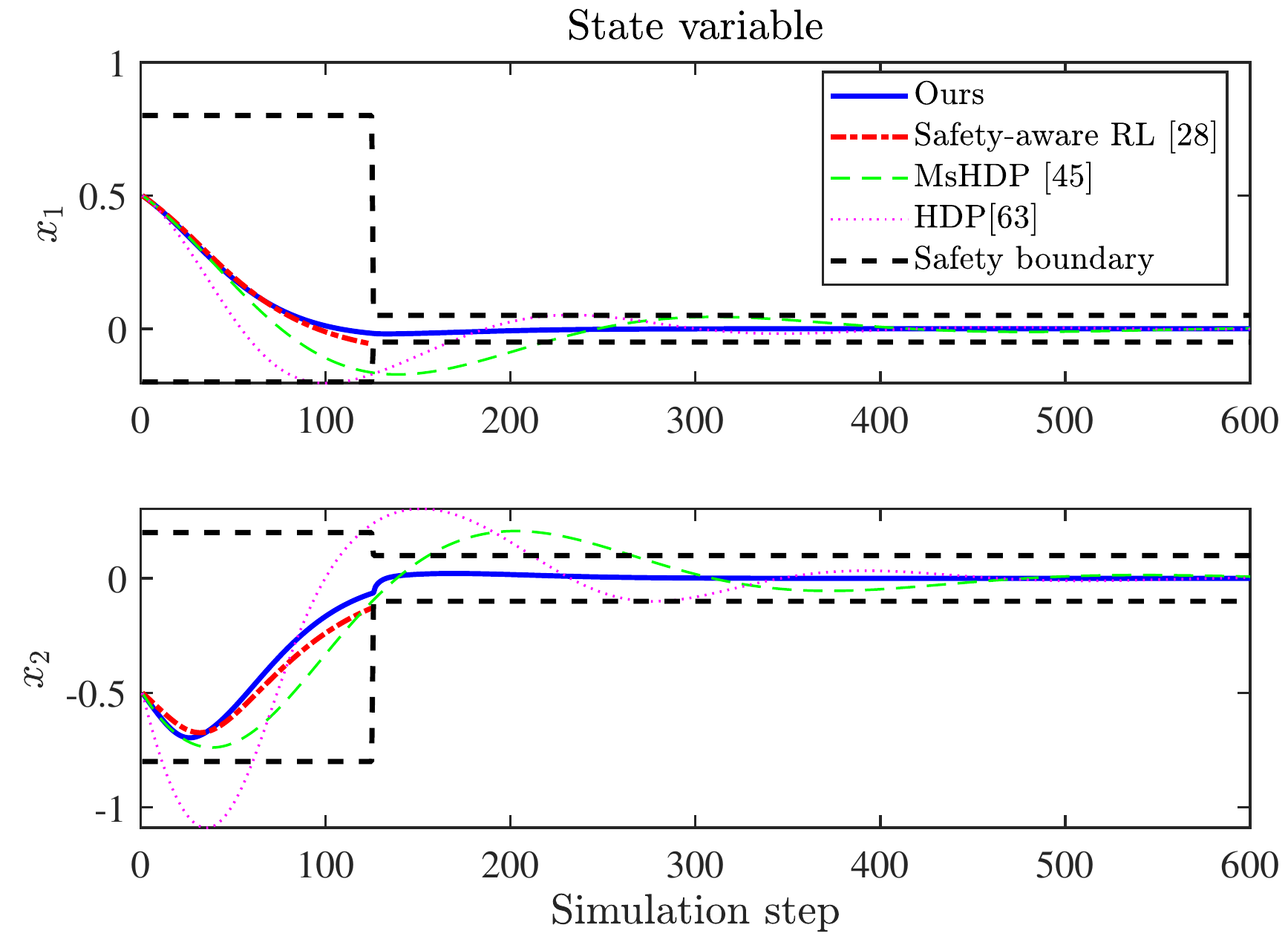}
			\caption{Simulation results for control of Van der Pol oscillator: The state variables (left panel) and the stage cost (right panel) with our approach and adopted comparative algorithms. {\color{black}In each training of safety-aware RL~\cite{yang2019safety}, DDPG-CS, and SAC-CS,  the control (learning) safety was not fulfilled, and the learning process was terminated when the size of state constraint was suddenly reduced. However, our approach can adapt to the constraint variation due to the adopted control policy structure and  multi-step policy evaluation mechanism.} }%
			\label{fig:com-cost-van}
		\end{figure*}
		\begin{table}[ht]
			\centering \caption{Comparisons in terms of safety rate in Van der Pol oscillator.}
			\label{tab:Tab_com3}
			\vskip 0.1cm
			\scalebox{0.75}{
				\begin{tabular}{@{}ccc@{}}
					\toprule
					Approach          & Ours &  Safety-aware RL~\cite{yang2019safety}\\
					\midrule
					%
					Time-invariant constraint            & 100\% & 100\%          \\ 
					Time-varying constraint             & 100\% & $\leq$ 50\%   \\ 
					\bottomrule
			\end{tabular}}
		\end{table}
		{\color{black} We compared our approach with two classic control methods, i.e., heuristic dynamic programming (HDP)~\cite{zhang2008novel}, multi-step heuristic dynamic programming (MsHDP)~\cite{luo2017output} and three safe RL approaches, e.g., safety-aware RL~\cite{yang2019safety}, DDPG-CS and SAC-CS. The control parameters of the proposed safe RL and the comparative approaches in~\cite{yang2019safety},~\cite{luo2017output}, and~\cite{zhang2008novel} were set similarly. In  DDPG-CS and SAC-CS, the cost function was reshaped with the same barrier functions adopted in the paper, and all the training parameters are fine-tuned according to~\cite{lillicrap2015continuous} and~\cite{haarnoja2018soft} respectively. The simulation results in Fig.~\ref{fig:com-cost-van} and Table~\ref{tab:Tab_com3} show that our approach can cope with time-varying state constraints in the control (learning) process while safety-aware RL in~\cite{yang2019safety}, DDPG-CS, and SAC-CS could fail due to the sudden change of constraints. The reason behind this is that the adopted safe RL approaches could not predict future changes of safety constraints and inform how to achieve safety by the actor-critic structure, hence prone to failing in abruptly changed environments.  MsHDP in~\cite{luo2017output} and HDP in~\cite{zhang2008novel}  can not guarantee safety constraint satisfaction. Moreover, our approach converged faster than that in~\cite{yang2019safety},~\cite{luo2017output}, and~\cite{zhang2008novel} (see Fig.~\ref{fig:com-cost-van}).}
	\end{document}